\documentclass{article}

\usepackage{iclr2027_conference,times}
\newif\ificlrpreprint
\newcommand{\iclrpreprint}{\iclrfinalcopy\iclrpreprinttrue}
\makeatletter
\g@addto@macro\@maketitle{\ificlrpreprint\lhead{Preprint}\fi}
\makeatother
\iclrpreprint

\usepackage{amsmath}
\usepackage{amssymb}
\usepackage{amsthm}
\newtheorem{proposition}{Proposition}

\usepackage{booktabs}
\usepackage{tabularx}
\usepackage{graphicx}
\usepackage{subcaption}
\usepackage{xcolor}
\usepackage{hyperref}
\usepackage{pdflscape}
\usepackage{placeins}

\usepackage{url}
\usepackage{hyperxmp}
\hypersetup{
  keeppdfinfo,  %
  pdftitle={High-Dimensional Simulation-Based Inference in Latent Spaces},
  pdfkeywords={latent generative models, simulation-based inference, amortized Bayesian inference, neural posterior estimation},
  pdfsubject={cs.LG, stat.ML},
  pdfcopyright={Licensed under the Creative Commons Attribution 4.0 International License (CC BY 4.0).},
  pdflicenseurl={https://creativecommons.org/licenses/by/4.0/},
}
\ificlrfinal
\hypersetup{pdfauthor={Lars K\"uhmichel, Stefan T. Radev, Bhanu Prasanna Koppolu, Masoumeh Davoudi, Jerry M. Huang, Paul-Christian B\"urkner}}
\else
\hypersetup{pdfauthor={Anonymous authors}}  %
\fi

\newif\ifdraft
\draftfalse
\ifdraft
  \newcommand{\todo}[1]{{\color{red}\textbf{[TODO: #1]}}}
  \newcommand{\scaffold}[1]{{\color{blue}#1}}
\else
  \newcommand{\todo}[1]{}
  \newcommand{\scaffold}[1]{}
\fi
\newcommand{\note}[1]{}

\newcommand{\btheta}{\boldsymbol{\theta}}  %
\newcommand{\bx}{\mathbf{x}}
\newcommand{\latent}{\boldsymbol{\vartheta}}  %
\newcommand{\bz}{\mathbf{z}}      %
\newcommand{\enc}{\mathcal{E}}    %
\newcommand{\dec}{\mathcal{D}}    %

\newcommand{\KL}{\operatorname{KL}}
\newcommand{\E}{\mathbb{E}}
\newcommand{\MI}{I}                        %

\newcommand{\msPerKdataFM}{20099}      %
\newcommand{\msPerKlatFM}{322}         %

\newcommand{\covErrFieldsFM}{0.071}       %
\newcommand{\covErrFieldsLatFM}{0.076}    %
\newcommand{\covErrFieldsFloor}{0.064}    %
\newcommand{\specCutFields}{13}           %
\newcommand{\specKeepHiFieldsLatFM}{9}    %
\newcommand{\specShareHiFields}{0.2}      %

\newcommand{\aeReconFashion}{0.008}    %
\newcommand{\aeReconToy}{0.0021}  %
\newcommand{\aeReconFashionOneK}{0.028}  %
\newcommand{\aeReconFashionTenK}{0.011}  %

\newcommand{\stepMsData}{45.6}  %
\newcommand{\stepMsLatent}{11.9}  %
\newcommand{\epochsData}{1{,}010}  %
\newcommand{\epochsLatent}{2{,}899}  %

\newcommand{\pretrainShareImage}{25\%}  %
\newcommand{\breakEvenFashionFM}{$2.7\times10^{5}$}  %

\newcommand{\flopEvalData}{2.1}        %
\newcommand{\flopEvalLatent}{0.018}    %
\newcommand{\decoderFlopShare}{5--14}  %

\newcommand{\speedupSat}{19}  %
\newcommand{\speedupToyCF}{2.3}  %
\newcommand{\speedupToyFM}{1.8}  %
\newcommand{\speedupToyDM}{3.6}  %

\newcommand{\codeDimToyKeight}{$16$}  %
\newcommand{\hCodeGivenThetaToyKeight}{$-46.4 \pm 3.7$}  %
\newcommand{\hCodeToyKeight}{$-6.6 \pm 4.3$}  %
\newcommand{\infoCodeToyKeight}{$39.8 \pm 0.7$}  %
\newcommand{\infoCodePerDimToyKeight}{$2.49 \pm 0.04$}  %
\newcommand{\aeMseToyKeight}{--}  %
\newcommand{\reconNllToyKeight}{--}  %
\newcommand{\hThetaToyKeight}{--}  %
\newcommand{\decoderTermToyKeight}{--}  %
\newcommand{\infoThetaXToyKeight}{$7.58$}  %
\newcommand{\codeTermLinearToyKeight}{$0$}  %
\newcommand{\linearKeptToyKeight}{$100\,\%$}  %
\newcommand{\genNllCFToyKeight}{$-14.2 \pm 4.1$}  %
\newcommand{\genNllFMToyKeight}{$-12.0 \pm 4.2$}  %
\newcommand{\cstCodeCFToyKeight}{$0.50 \pm 0.01$}  %
\newcommand{\cstCodeFMToyKeight}{$0.50 \pm 0.02$}  %
\newcommand{\cstCodeDMToyKeight}{$0.51$}  %
\newcommand{\cstCodeNullToyKeight}{$0.50$}  %
\newcommand{\codeDimFashion}{$512$}
\newcommand{\hCodeGivenThetaFashion}{$-315.0 \pm 2.8$}
\newcommand{\hCodeFashion}{$573.2 \pm 0.9$}
\newcommand{\infoCodeFashion}{$888.2 \pm 3.5$}
\newcommand{\infoCodePerDimFashion}{$1.73 \pm 0.01$}
\newcommand{\aeMseFashion}{$0.0006$}
\newcommand{\reconNllFashion}{$-3450.4 \pm 2.0$}
\newcommand{\hThetaFashion}{--}
\newcommand{\decoderTermFashion}{--}
\newcommand{\infoThetaXFashion}{--}
\newcommand{\codeTermLinearFashion}{--}
\newcommand{\linearKeptFashion}{--}
\newcommand{\genNllCFFashion}{--}
\newcommand{\genNllFMFashion}{$262.4 \pm 1.1$}
\newcommand{\cstCodeCFFashion}{--}
\newcommand{\cstCodeFMFashion}{$0.52$}
\newcommand{\cstCodeDMFashion}{$0.54$}  %
\newcommand{\cstCodeNullFashion}{$0.50$}
\newcommand{\codeDimFields}{$512$}
\newcommand{\hCodeGivenThetaFields}{$-891.1 \pm 1.5$}
\newcommand{\hCodeFields}{$646.5 \pm 1.0$}
\newcommand{\infoCodeFields}{$1537.6 \pm 0.9$}
\newcommand{\infoCodePerDimFields}{$3.00$}
\newcommand{\aeMseFields}{$0.0186$}
\newcommand{\reconNllFields}{$-2797.0 \pm 1.2$}
\newcommand{\hThetaFields}{$-2181.2$}
\newcommand{\decoderTermFields}{$921.8 \pm 2.0$}
\newcommand{\infoThetaXFields}{$4.45$}
\newcommand{\codeTermLinearFields}{--}
\newcommand{\linearKeptFields}{--}
\newcommand{\genNllCFFields}{--}
\newcommand{\genNllFMFields}{$620.5 \pm 1.6$}
\newcommand{\cstCodeCFFields}{--}
\newcommand{\cstCodeFMFields}{$0.50$}
\newcommand{\cstCodeDMFields}{$0.51 \pm 0.01$}  %
\newcommand{\cstCodeNullFields}{$0.50$}

\newcommand{\cstToyExact}{0.50}  %

\newcommand{\ftNrmseJoint}{0.065}   %
\newcommand{\ftNrmseFrozen}{0.033}  %
\newcommand{\ftNrmseKLmid}{0.049}   %
\newcommand{\ftNrmseKLhigh}{0.104}  %
\newcommand{\ftStdJoint}{0.007}     %
\newcommand{\ftReconJoint}{1.5}     %
\newcommand{\ftStdKLmid}{0.32}      %
\newcommand{\ftStdKLhigh}{0.87}     %
\newcommand{\ftReconKLmid}{7.0}     %
\newcommand{\ftReconKLhigh}{13.6}   %

\newcommand{\specErrFieldsData}{0.012}  %
\newcommand{\specErrFieldsLat}{0.46}    %
\newcommand{\specErrFieldsFloor}{0.004} %
\newcommand{\covErrFieldsDM}{0.082}     %
\newcommand{\covErrFieldsLatDM}{0.097}  %

\newcommand{\cnnFieldsLat}{1.00}     %
\newcommand{\cnnFieldsRecon}{1.00}   %
\newcommand{\cnnSatLat}{0.98}        %
\newcommand{\cnnSatRecon}{0.99}      %

\title{High-Dimensional Simulation-Based\\ Inference in Latent Spaces}

\author{Lars K\"uhmichel$^{1}$, Stefan T. Radev$^{2}$, Bhanu Prasanna Koppolu$^{1}$, Masoumeh Davoudi$^{1}$,\\
\bf Jerry M. Huang$^{2}$ \& Paul-Christian B\"urkner$^{1}$ \\
$^{1}$Department of Statistics, TU Dortmund University, Germany \\
$^{2}$Department of Cognitive Science, Rensselaer Polytechnic Institute \\
Correspondence: \texttt{larskuedev@gmail.com}}

\begin{document}

\maketitle

\begin{abstract}
Neural simulation-based inference (SBI) has been widely successful in inferring a relatively small number of interpretable parameters from potentially high-dimensional observations, such as images or time series. Accordingly, representation learning in SBI has focused almost exclusively on compressing the observations used to condition the posterior. More recently, however, SBI has begun to target increasingly high-dimensional parameter spaces, raising the complementary question of whether the inference target itself should be compressed. Our answer is a practical merger of SBI and latent generative modeling, which learns a low-dimensional representation of the simulator parameters, performs posterior inference directly in this latent space, and maps posterior samples back to the original parameter space. We characterize the conditions under which latent-space inference recovers the desired target posterior and systematically study its empirical trade-offs. Across four case studies and three generative families, we compare latent and standard estimators while controlling for network capacity, regularization, optimization, and training compute. At matched training compute, latent-space inference achieves accuracy and marginal calibration comparable to direct target-space inference while sampling up to more than an order of magnitude faster.

\end{abstract}

\section{Introduction}

Simulation-based inference \citep[SBI;][]{cranmer2020frontier} has emerged as a powerful paradigm for estimating the parameters of mathematical models from highly complex \citep{dax2025real} or large-scale \citep{von2026scaling} data. In \textit{amortized} SBI, a neural network is first trained on labeled simulations. Once trained, the network processes incoming data in inference mode, eventually amortizing the simulation and training costs. Most SBI applications have traditionally considered fully Bayesian inference over relatively low-dimensional parameters $\btheta$ even when the observation $\bx$ is high-dimensional. However, recent research has begun to shift this regime toward inference over very high-dimensional parameter spaces \citep[see][for an overview]{arruda2025diffusion}.

SBI and Bayesian inverse problems have confronted high dimensionality on opposite sides of the inference problem. SBI has focused on compressing high-dimensional \textit{observations} through domain-specific preprocessing \citep{papamakarios2016fast}, summary networks \citep{radev2020bayesflow}, or both \citep{dax2025real}. Bayesian inverse problems, by contrast, often involve extremely high-dimensional \textit{parameters} (e.g., the coefficient fields of PDEs) and the literature abounds with methods that reduce the effective dimension of the posterior, for instance, through likelihood-informed subspaces, low-dimensional couplings, and autoencoders \citep{cui2014likelihood,cui2016scalable,zahm2022certified,spantini2018inference,baptista2022gradient,lan2022scaling}. These methods are largely tailored to particular inverse problems or algorithms and are thus not portable to amortized SBI out of the box.

At the same time, latent generative modeling has made learning and sampling in compressed target spaces standard across autoregressive, diffusion, flow-matching, and consistency models \citep{oord2017vqvae,rombach2022ldm,dao2023lfm,luo2023lcm}. Here, however, representation learning is designed primarily for efficient and faithful \emph{synthesis}, not for preserving statistical accuracy and uncertainty. In the present work, we bring these research directions together and show how to compress high-dimensional posterior targets for amortized SBI while preserving the information required for accurate and calibrated Bayesian inference.

\begin{figure}[t]
    \centering
    \includegraphics[width=\linewidth]{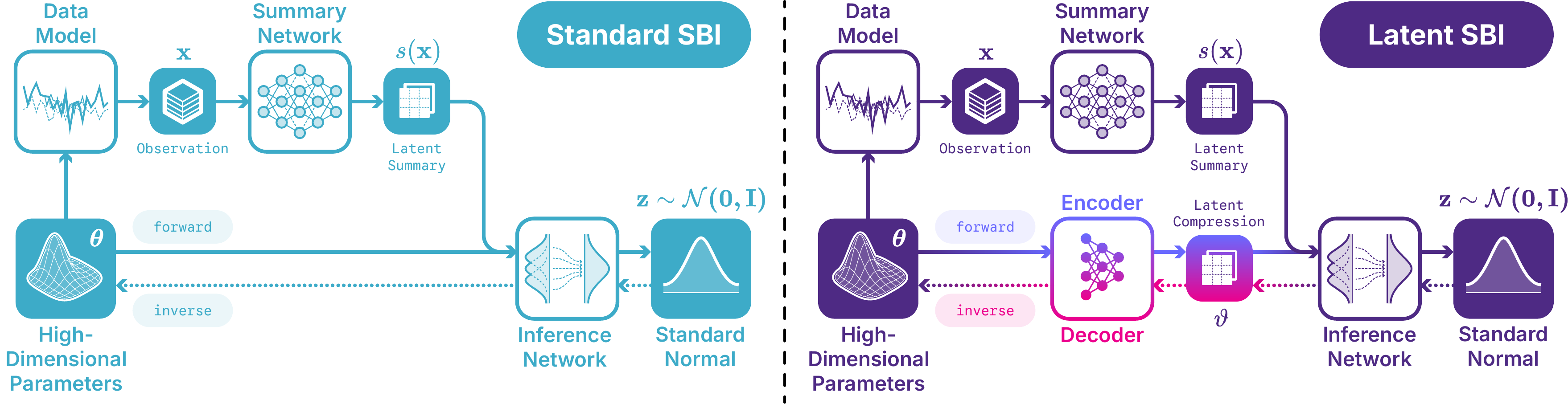}
    \caption{\textbf{Side-by-side comparison between standard (\textit{left}) and latent (\textit{right}) simulation-based inference (SBI) approaches.} In latent SBI, high-dimensional inference targets are compressed into a low-dimensional representation using an encoder. This representation is later used for target reconstruction. The encoder maps parameters $\btheta$ to a code $\latent = \enc(\btheta)$; given the summary $s(\bx)$, the inference network transports base noise $\bz \sim \mathcal{N}(\mathbf{0}, I)$ to $\latent$, and the decoder returns $\btheta = \dec(\latent)$.}
    \label{fig:architecture}
\end{figure}

Our proposed approach for \textit{latent} amortized SBI (\autoref{fig:architecture}) (1) learns an information-maximizing autoencoder over the parameter space of the simulator, (2) freezes its representation, (3) learns a conditional model to approximate the posterior over latent codes, and (4) decodes posterior samples back into the original parameter space. 
We provide practical mathematical conditions under which the proposed approach provides correct results and stable settings that generalize across different settings. In several simulated and real-world case studies, we numerically compared our latent-space estimators against otherwise matched target-space estimators, controlling network capacity, regularization, optimization, and training compute. Our results deliver fast sampling for high-dimensional SBI that maintains accuracy and marginal calibration at equal training compute with standard methods.

\section{Method}
\subsection{Simulation-Based Inference with High-Dimensional Targets}
\label{sec:setting}

Our goal is to train generative networks to perform efficient simulation-based inference \citep{cranmer2020frontier, deistler2025simulation, arruda2025diffusion}. Given samples from a joint distribution $p(\btheta, \bx) = p(\btheta)\,p(\bx \mid \btheta)$ of parameters $\btheta$ and observables $\bx$, we want to approximate the posterior $p(\btheta \mid \bx) \propto p(\btheta)\,p(\bx \mid \btheta)$. As is typical in SBI, we assume that evaluating the density of the joint distribution is intractable, but samples are available. We therefore use samples $(\btheta_n, \bx_n)$ from the joint distribution to train an \textit{amortized distribution} $q_\phi(\btheta \mid \bx)$ that can draw approximate posterior samples for any $\bx$.

When working with complex data $\bx$ (e.g., time series, images), a summary network $s$ typically compresses $\bx$ into a lower-dimensional summary $s(\bx)$ that conditions $q_\phi$ in place of $\bx$ \citep{radev2020bayesflow, chen2021sufficient}. This process is now commonplace in SBI, as the dimensionality of $\bx$ is treated merely as an obstacle to infer the parameters $\btheta$, the ultimate objects of interest. Since the coordinates of $\btheta$ carry the inferential meaning, the parameter space itself is typically left intact.

However, several applications deviate from this schema. For instance, in Bayesian denoising \citep{schmitt2024cmpe}, the ``parameters'' $\btheta$ are the clean version of the corrupted data $\bx$ and thus have the same dimensionality. In neural likelihood estimation \citep[NLE;][]{papamakarios2019sequential}, the inference target is instead an emulator $q_\phi(\bx \mid \btheta)$, so the learned distribution is over the potentially high-dimensional $\bx$ whose condition $\btheta$ may not require any pre-processing. Finally, some of the most challenging applications of amortized inference involve recovering a high-dimensional image from a series of high-dimensional measurements, as in ultrasound imaging \citep{orozco2025aspire}. In that case, $\btheta$ as a whole is only interpretable through the lens of a domain expert.

\subsection{Two-Stage Latent Inference}
\label{sec:ambient}
\label{sec:two-stage}

In all of the above settings, the \textit{ambient dimensionality} of $\btheta$ may overstate the complexity of the inference problem. Although $\btheta \in \mathbb{R}^D$ has $D$ coordinates, the posterior may vary along far fewer independent directions due to correlations or other constraints. We refer to this concept as the \textit{effective dimension} $d_{\mathrm{eff}}$ \citep{levina2004maximum, pope2021intrinsic}, the number of $\btheta$-directions that must be represented for posterior inference. Even when every coordinate of $\btheta$ is of scientific interest and must ultimately be recovered, inference can still be performed on a lower-dimensional representation.

Posterior dependencies can manifest themselves in relationships more complex than pairwise linear dependencies. To capture them, we define $d_{\mathrm{eff}}$ from an information-theoretic lens \citep{li1991sliced, baptista2022gradient} as
\begin{equation}
d_{\mathrm{eff}}(\varepsilon)=\min\left\{k\in\{1,\dots,D\}:\exists\,T_k:\mathbb{R}^{D}\to\mathbb{R}^{k}\ \text{s.t.}\ I\!\left(\btheta;\bx\mid T_k(\btheta)\right)\leq\varepsilon\right\},\qquad \varepsilon\geq 0,
\end{equation}
where $I$ is the conditional mutual information and $T_k$ is a map to $k$ dimensions.

To learn such a map, we introduce an encoder $\enc$ that maps the high-dimensional target $\btheta$ to a lower-dimensional latent representation, $\latent = \enc(\btheta)$. We train the generative network $q$ to approximate the posterior over this representation. A decoder $\dec$ then maps posterior samples in the latent space $\latent \sim q_{\phi}(\latent \mid \bx)$ back to the original parameter space, $\btheta = \dec(\latent)$. The encoder sees only $\btheta$, so without knowledge of $\bx$, it can at best match the intrinsic dimension of $p(\btheta)$, an upper bound on $d_{\mathrm{eff}}(\varepsilon)$. In this way, the generative network only needs to solve the inference problem in the ``effective subspace'', while the decoder restores samples in the original, interpretable parameter space.

Our latent SBI approach proceeds in two stages, inspired by \citet{rombach2022ldm} and our empirical results. The \textit{first stage} trains an InfoVAE \citep{zhao2019infovae}, a generalization of the classical VAE, on samples $\btheta \sim p(\btheta)$ alone. The encoder defines a Gaussian $q_\xi(\latent \mid \btheta)$ with learned mean and variance, the decoder is deterministic, and the objective is the reconstruction error $\E_{\btheta, \latent}\|\btheta - \dec(\latent)\|^2$ plus a KL penalty that pulls each code distribution $q_\xi(\latent \mid \btheta)$ toward a standard normal prior and a maximum mean discrepancy that pulls their aggregate toward it, weighted by $1-\alpha$ and $\alpha+\lambda-1$ (\autoref{app:config}).
The autoencoder is then frozen.
The \textit{second stage} trains the amortized posterior $q_\phi(\latent \mid \bx)$ on codes $\latent \sim q_\xi(\latent \mid \btheta)$ of the training pairs. No gradient reaches the encoder, so the second stage cannot reshape the code it is given. Fine-tuning the encoder jointly with the posterior network made every variant in our ablation less accurate than a frozen control (\autoref{app:ablations}). See \autoref{app:config} for more details. Crucially, the autoencoder ``unrestricts'' the choice of posterior network. To demonstrate this flexibility, we use three common architectures in our experiments: flow matching \citep{lipman2023flow, liu2023rectified}, diffusion \citep{ho2020ddpm, song2021sde, karras2022edm}, and normalizing flows \citep{dinh2017realnvp, papamakarios2021normalizing}, to account for different speeds and resolution of the generative processes.

\subsection{Error Decomposition of Latent Compression}
\label{sec:theory}

Compared to target-space SBI, latent SBI adds two sources of error: the code can discard information that $\bx$ carries about $\btheta$, and the decoder can fail to invert the encoder. We separate the errors that each stage controls to identify which stage limits the accuracy of the latent posterior.

In \autoref{prop:decomposition}, we show that the error of the joint posterior on $\latent$ and $\btheta$ with respect to the analytic posterior on $\btheta$ is bounded by the sum of a posterior, a code, and a decoder term. In \autoref{prop:objective}, we show that the first-stage maximizer at fixed mutual information $\MI(\btheta; \latent)$ removes the decoder term.

The second stage fits $q_\phi(\latent \mid \bx)$ to the latent posterior $q_\xi(\latent \mid \bx)$ induced by the encoder, and we treat the decoder as the Gaussian likelihood $q_\psi(\btheta \mid \latent) = \mathcal{N}(\dec(\latent), \sigma^2 I)$ with fixed $\sigma > 0$. The latent approximation to the posterior is then $q_{\phi, \psi}(\btheta \mid \bx) = \int q_\psi(\btheta \mid \latent)\, q_\phi(\latent \mid \bx)\, d\latent$.

\vspace{0.1cm}

\begin{proposition}
\label{prop:decomposition}
Let $\KL$ denote the Kullback--Leibler divergence and $\MI$ the mutual information
under $q_\xi$. Assume $\MI(\btheta; \bx) < \infty$ and that all divergences below are finite. Then
\begin{equation}
\label{eq:decomposition}
\begin{split}
\E_{\bx} \KL\big(q_\xi(\btheta, \latent \mid \bx) \,\|\, q_\phi(\latent \mid \bx)\, q_\psi(\btheta \mid \latent)\big)
&= \underbrace{\E_{\bx} \KL\big(q_\xi(\latent \mid \bx) \,\|\, q_\phi(\latent \mid \bx)\big)}_{\text{posterior}}
+ \underbrace{\MI(\btheta; \bx) - \MI(\latent; \bx)}_{\text{code}} \\
&\quad + \underbrace{\E_{\latent} \KL\big(q_\xi(\btheta \mid \latent) \,\|\, q_\psi(\btheta \mid \latent)\big)}_{\text{decoder}},
\end{split}
\end{equation}
and $\E_{\bx} \KL\big(p(\btheta \mid \bx) \,\|\, q_{\phi, \psi}(\btheta \mid \bx)\big)$ is at most
the left-hand side.
\end{proposition}

The posterior term is the amortization error of SBI in code space and the only term that the second stage can change (proof in \autoref{app:proofs}). The code term is the information about $\btheta$ that $\bx$ carries and the code discards, and the decoder term measures how far the decoder is from inverting the encoder.

The first stage controls the other two terms. At the optimum of its objective, the decoder inverts the encoder:

\begin{proposition}
\label{prop:objective}
The first-stage objective of \autoref{sec:two-stage} is an instance of the InfoVAE
objective \citep{zhao2019infovae} with $\alpha < 1$ and $\lambda > 0$. For sufficiently
flexible encoder and decoder families, the maximizer at fixed $\MI(\btheta; \latent)$
makes the decoder term of \eqref{eq:decomposition} vanish, and the maximal value depends on
$q_\xi(\latent \mid \btheta)$ only through $\MI(\btheta; \latent)$.
\end{proposition}

The first stage also targets the code term, through the reconstruction error. The code term equals $\MI(\btheta; \bx \mid \latent)$, the quantity that defines the effective dimension in \autoref{sec:ambient}, so if the code dimension is at least $d_{\mathrm{eff}}(\varepsilon)$, some code keeps the code term below $\varepsilon$, in analogy to a sufficient summary of $\bx$.

Whether the first stage finds such a code is a separate question, because it never sees $\bx$. When the code cannot keep all of $\btheta$, the encoder keeps the directions that reduce the reconstruction error most, for a linear code those of largest prior variance, whether or not $\bx$ informs them. We therefore ablate the compression ratio (\autoref{fig:toy-example-compressions}).

\section{Related Work}
\label{sec:related-work}

Throughout SBI, dimensionality reduction has primarily been applied to the \textit{observation side} of the inference problem via domain-specific preprocessing or learned summary representations \citep[see][]{deistler2025simulation, arruda2025diffusion}. Its theoretical justification is sufficiency: a representation may be reduced if it preserves the information for the posterior without inferential loss. By contrast, the inference target is typically retained in full. This asymmetry motivated the question of whether one can apply an analogous reduction to the target for scalable amortized inference.

Bayesian inverse problems provide an important precedent for such \textit{target-side} reduction, using, for instance, likelihood-informed subspaces, low-dimensional parameterizations, and certified reductions to concentrate inference in lower-dimensional spaces \citep{cui2014likelihood, cui2016scalable, zahm2022certified, spantini2018inference, baptista2022gradient, lan2022scaling}. These constructions, however, are generally constructed for a particular forward model, likelihood, or sampling algorithm and therefore do not directly provide a general parameter compression mechanism for amortized SBI.

A distinct line of work uses dimensionality reduction to simplify the modeling of high-dimensional unstructured data. For instance, in high-resolution image generation, a common strategy is to encode observations into a lower-dimensional latent representation, learn a generative model in that latent space, and decode samples back to the original space. This two-stage recipe is well established \citep{vahdat2021lsgm, rombach2022ldm, dao2023lfm, luo2023lcm}, alongside discrete representation learning \citep{oord2017vqvae} and densities fitted \textit{post hoc} to frozen representations \citep{ghosh2020rae}. However, it has not yet found its way into SBI, where statistical accuracy instead of sample quality is the ultimate metric of success.

Differently, some of these ideas have been imported into Bayesian inverse problems: deep generative priors reparameterize unknown fields and infer in a reduced latent space \citep{patel2022priors}, whereas guided samplers condition pretrained generative models on observations in pixel space \citep{chung2024decomposed, kawar2022denoising}, latent space \citep{rout2024beyond, song2024solving}, or across broader model classes \citep{venkatraman2025outsourced}. While these methods compress high-dimensional targets into lower-dimensional representations, they generally rely on an explicit likelihood, gradients, or a forward operator, and require a new guided-sampling procedure for each observation. In our work, we aim to establish whether learned target compression can systematically improve amortized SBI without compromising statistical accuracy.

\section{Experiments}
\label{sec:experiments}
\subsection{Case Studies}
\label{sec:simulators}

We evaluate on four case studies with increasing complexity, from a fully analytic (yet challenging) target to practical Bayesian denoising tasks (details in \autoref{app:simulators}).
All of these tasks are concerned with parameters of much higher dimensions than those encountered in typical SBI applications and benchmarks, which tend to focus on settings with $\mathrm{dim}(\btheta) < 30$ \citep{arruda2025diffusion}.

\paragraph{Correlated Gaussian} The parameters of interest in our first study is the covariance matrix $\Sigma = \operatorname{blockdiag}(\sigma_1^2 \mathbf{1}\mathbf{1}^\top, \dots, \sigma_k^2 \mathbf{1}\mathbf{1}^\top)$ of a $64$-dimensional Gaussian with $k = 8$ blocks, and $\btheta$ collects its $2{,}080$ upper-triangular entries. The observation $\bx$ is the per-coordinate second moment of $32$ draws from $\mathcal{N}(\mathbf{0}, \Sigma)$, which is sufficient for the block scales. Since $\btheta$ is a deterministic function of the $k$ scales, it has $2{,}080$ coordinates but only $k$ degrees of freedom. $1{,}792$ of its entries are structurally zero, and the element-wise metrics of \autoref{sec:metrics} include them, whereas \autoref{fig:results-toy-exact} compares the block variances. The posterior factorizes over blocks into one-dimensional densities from which we draw exact samples, so that we can check every method directly against the exact posterior (\autoref{app:simulators}).

\paragraph{Gaussian random fields} In our second study, the parameters comprise a field $\btheta \in \mathbb{R}^{32 \times 32}$ with a power-law spectrum, simulated on the FFT grid \citep{lang2011grf}. The two-dimensional observable $\bx = (\log\sigma, \alpha)$ contains the spectrum's log-amplitude and exponent, the latter controlling the field's smoothness. 
A power-law spectrum concentrates the variance in a few low frequencies, so the field is highly redundant.
Because $\alpha$ is drawn per sample, the redundancy also varies from one field to the next, making the problem a very challenging task for SBI methods, as shown in \citet{arruda2025diffusion}.
Given $\bx$, the field is an exact Gaussian process, so the posterior is Gaussian with an analytically known covariance, against which we compare the sampled posteriors directly.

\paragraph{Fashion-MNIST deblurring} Deblurring recovers a Fashion-MNIST image $\btheta \in [0, 1]^{32 \times 32}$ from a blurred, noisy copy $\bx$ of the same shape, generated by a simulated camera with shot noise (\autoref{app:simulators}). This is the Bayesian denoising task of \citet{schmitt2024cmpe}. Blur removes high-frequency detail, so many images are consistent with one blurred observation. The redundancy of $\btheta$ is that of natural images, whose intrinsic dimension is far below their pixel count \citep{pope2021intrinsic}.

\paragraph{Map to satellite inference} Finally, we infer an RGB satellite image $\btheta \in [0, 1]^{256 \times 256 \times 3}$ from the map image $\bx$ of the same location and shape, taken from a corpus of aligned pairs \citep{sun2025cscmg}. Image-to-image translation has been used to generate maps from satellite images \citep{isola2017pix2pix}. In that direction, the posterior is almost a point mass, since the map is close to a deterministic function of the satellite image. Such a posterior concentrates near a lower-dimensional set, which makes it a challenging distribution for most generative families to represent. We therefore infer in the inverse direction, where the map leaves attributes such as roof material, vehicles, and season unconstrained. The redundancy in $\btheta$ is again that of images, but at a much higher resolution than Fashion-MNIST.

\subsection{General Setup}
\label{sec:protocol}

Across all case studies, we compare latent- and target-space variants at matched training compute and model size, while holding all other training details fixed. Compute is defined as wall-clock training time, excluding data generation and evaluation. A short timing run estimates time per training step, which determines the number of epochs allocated to each variant. We match the model size by counting the latent variant's frozen autoencoder together with its posterior and summary networks, because the autoencoder remains part of the inference model (\autoref{app:config}). 
Importantly, our latent training budget also includes autoencoder training; excluding this cost would amount to assuming a pretrained encoder. Optimizer, schedule, batch size, data volume, seeds, and numerical precision are otherwise identical. We implement all networks and training procedures in BayesFlow~2 \citep{kuhmichel2026bayesflow}; further details appear in \autoref{app:protocol}.

\begin{figure}[!t]
  \centering
  \includegraphics[width=\linewidth]{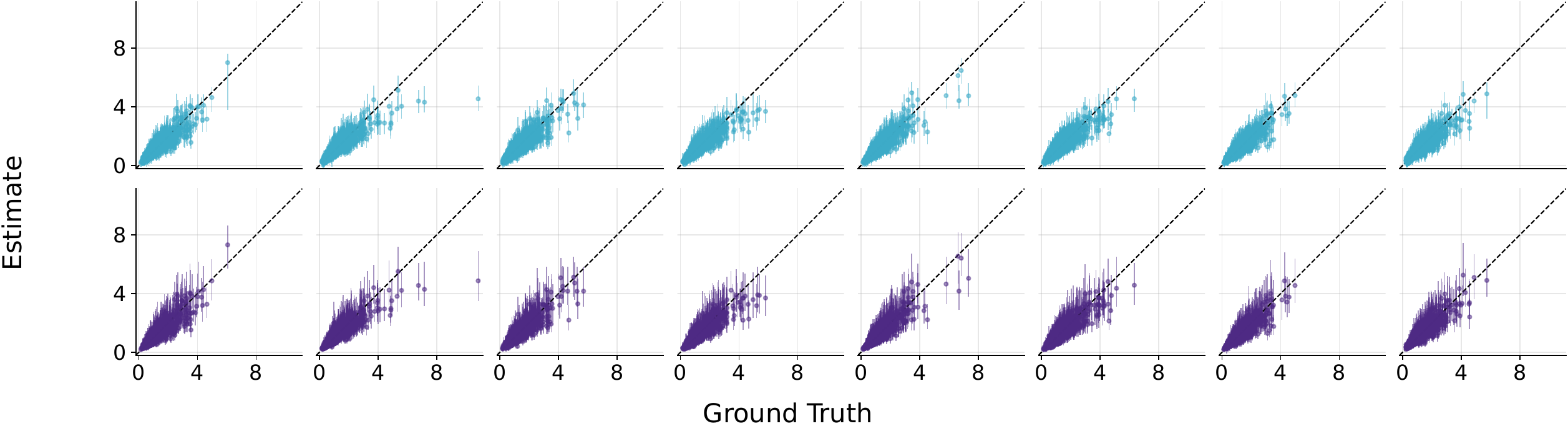}\\[6pt]
  \includegraphics[width=\linewidth]{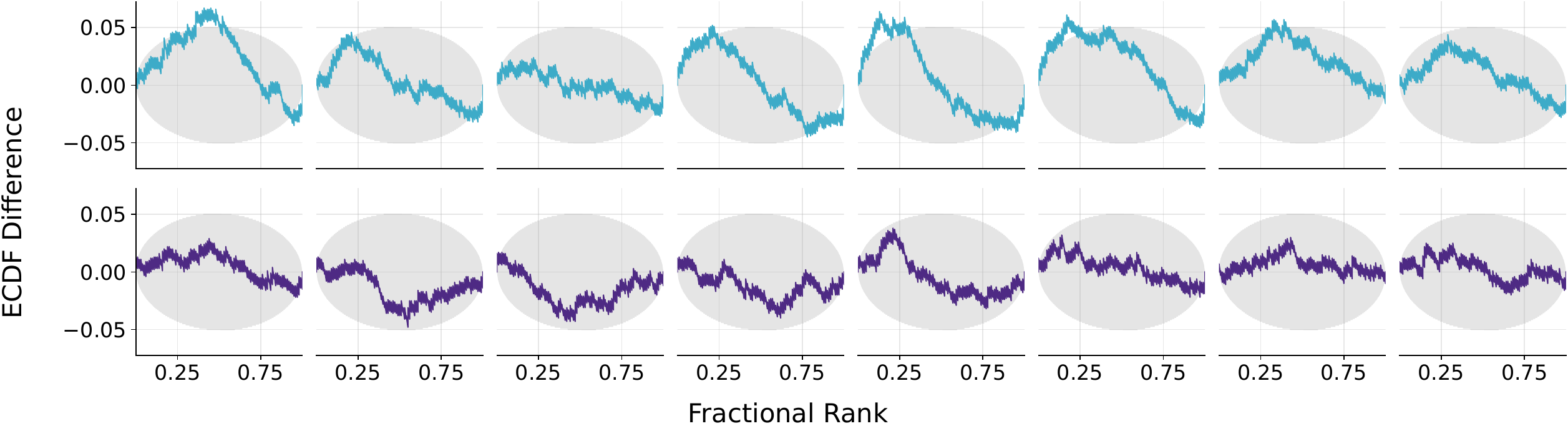}
  \caption{\textbf{Recovery and calibration on the correlated Gaussian.} Parameter recovery (\textit{top}) and calibration ECDFs \citep{sailynoja2022graphical} (\textit{bottom}) of standard SBI (\textit{blue}, first row of each) and latent SBI (\textit{purple}, second row of each) for the correlated Gaussian with flow matching, one column per block variance $\sigma_1^2, \dots, \sigma_8^2$. Latent SBI achieves notably better calibration than standard SBI, as no ECDF leaves the simultaneous confidence bands. A comparison against the exact posterior is in \autoref{app:additional}.}
  \label{fig:results-toy}
\end{figure}

\begin{figure}[!t]
  \centering
  \begin{subfigure}{0.49\linewidth}
    \includegraphics[width=\linewidth]{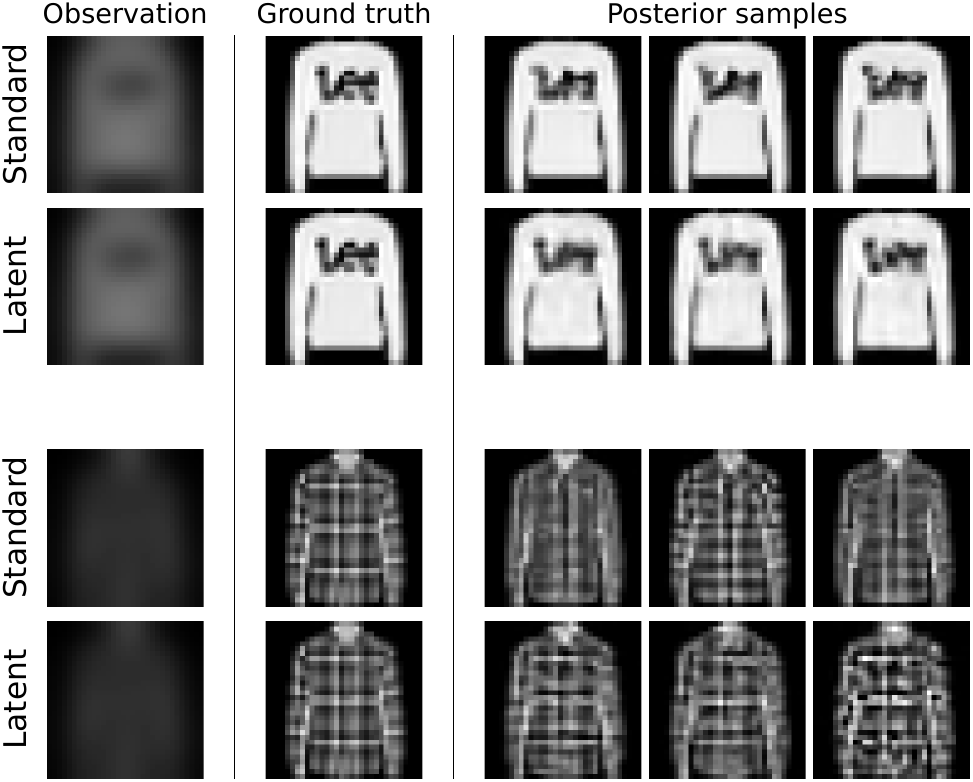}
    \caption{Fashion-MNIST deblurring}
    \label{fig:results-fashion}
  \end{subfigure}\hfill
  \begin{subfigure}{0.49\linewidth}
    \includegraphics[width=\linewidth]{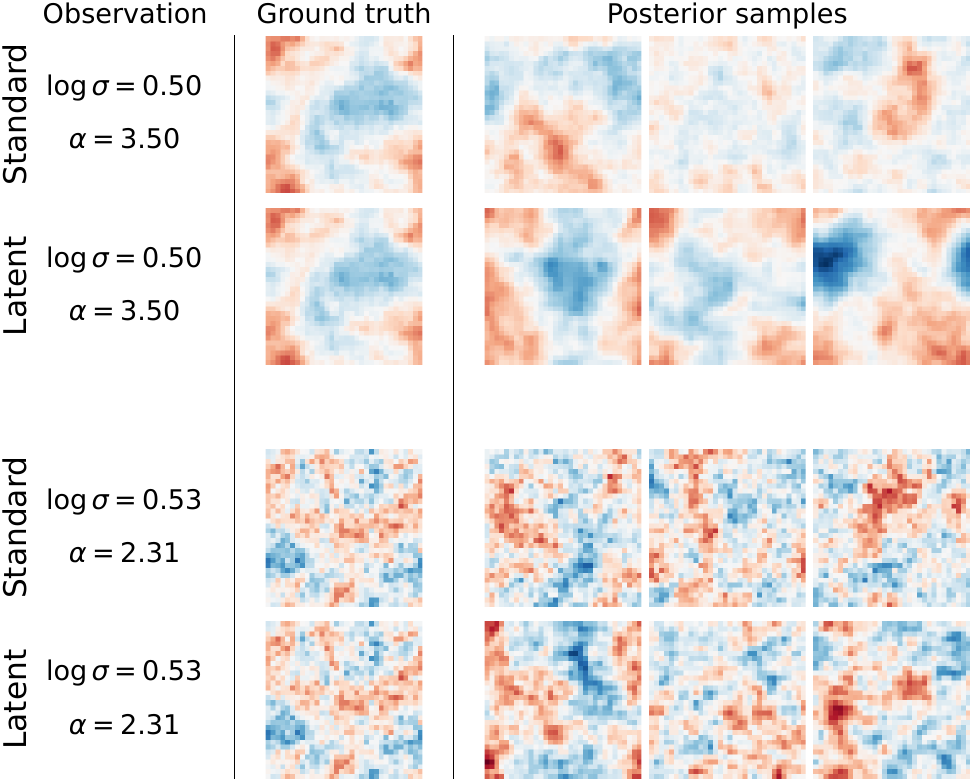}
    \caption{Gaussian random fields}
    \label{fig:results-fields}
  \end{subfigure}
  \caption{\textbf{Posterior draws with latent and standard flow matching for two conditions of the Bayesian denoising and GRF estimation case studies.} For
  the random fields, the posterior is the whole Gaussian process, so draws should match the
  amplitude and smoothness of the ground truth, not its pixel values. Numerical results are in \autoref{tab:results} and random conditions are in
  \autoref{app:per-seed}.}
  \label{fig:results-images}
\end{figure}

\begin{figure}[!t]
  \centering
  \includegraphics[width=\linewidth]{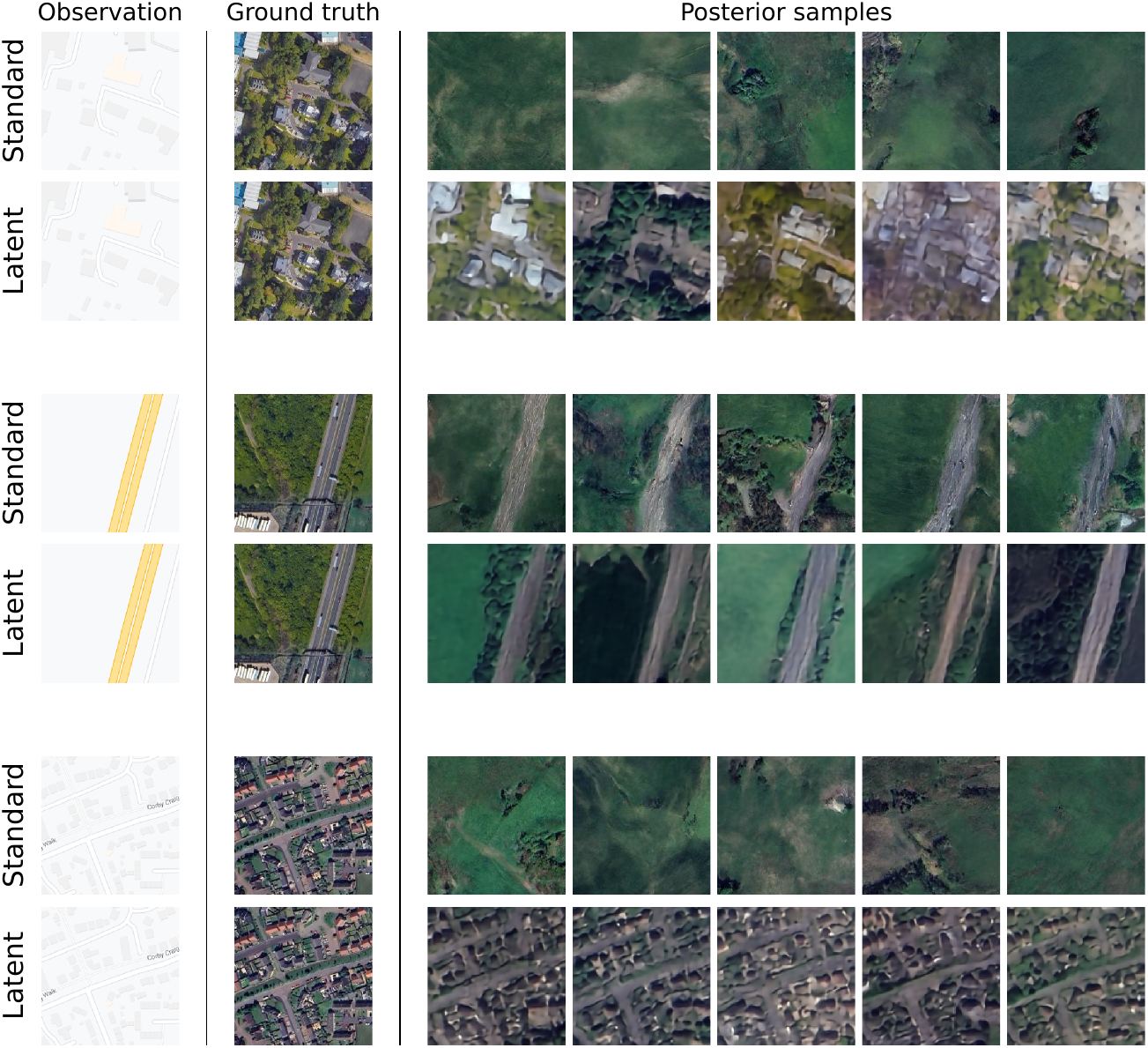}
  \caption{\textbf{Posterior draws for map to satellite with diffusion at $256 \times 256$ under a matched training budget.} Latent draws follow the street layout of the map, while target-space draws
  mostly show generic vegetation. Latent sampling is also \speedupSat$\times$ faster. Numerical results are in \autoref{tab:results} and random
  conditions are in \autoref{app:per-seed}.}
  \label{fig:results-map2sat}
\end{figure}

\subsection{Metrics}
\label{sec:metrics}

All metrics below are sample-based and computed in the target space to ensure comparability between parameter- and latent-space inference. No metric requires a tractable density. Further details are in \autoref{app:calibration} and \autoref{app:compute}.

\paragraph{Accuracy} We use the root mean square error (NRMSE) of the posterior mean, normalized by the inference targets’ value range. The range is one for images in $[0, 1]$ and is measured on the first $32$ test targets of the correlated Gaussian and the random fields. To score posterior accuracy and uncertainty together, we additionally report the continuous ranked probability score \citep[CRPS;][]{gneiting2007crps}, which is strictly proper for each coordinate, so neither a posterior that is too wide nor one that is too narrow can improve it. On the image case studies, we also report the structural similarity index \citep[SSIM;][]{wang2004ssim} between single posterior draws and the ground truth.

\paragraph{Calibration} At the marginal level, we report element-wise central-interval coverage at the $50\%$ and $90\%$ levels, which a calibrated posterior matches, and the calibration error \citep{kuhmichel2026bayesflow}, which summarizes the whole coverage curve: the median over target dimensions of the median absolute difference between empirical and nominal coverage across $20$ central credible levels from $0.5\%$ to $99.5\%$. At the joint level, we report the classifier two-sample test \citep[C2ST;][]{lopez2016revisiting}. Its accuracy is optimal at 0.5, where draws and true targets are indistinguishable. A higher value suggests the two are told apart more easily.

\paragraph{Sampling cost} We report the sampling speedup of the latent variant, $t_{\text{standard}} / t_{\text{latent}}$, where $t$ is the measured time per $1000$ posterior draws. Since sampling speed differs strongly between generative model families, we only compare standard and latent sampling within each family.

\subsection{Results}
\label{sec:results}

\begin{table}[t]
  \centering
  \caption{\textbf{Posterior quality and sampling cost of standard (target-space) and latent inference
  at an equal training budget.} We report one seed for the map to satellite case study and the mean $\pm$ standard error over three seeds otherwise.
  Definitions of the metrics are in \autoref{sec:metrics}.}
  \label{tab:results}
  \scriptsize
  \setlength{\tabcolsep}{3pt}
  \begin{tabular}{@{}ll rrr rrr r r@{}}
    \toprule
    & & \multicolumn{2}{c}{Accuracy} & \multicolumn{4}{c}{Calibration} & & Cost \\
    \cmidrule(lr){3-4} \cmidrule(lr){5-8} \cmidrule(lr){10-10}
    & & NRMSE\textsuperscript{$\ddagger$} $\downarrow$ & CRPS\textsuperscript{$\ddagger$} $\downarrow$ & Cal.\ err.\textsuperscript{$\ddagger$} $\downarrow$ & Cov$_{50}$\textsuperscript{$\ast$} & Cov$_{90}$\textsuperscript{$\ast$} & C2ST\textsuperscript{$\dagger$} & SSIM\textsuperscript{$\ddagger$} $\uparrow$ & Speedup $\uparrow$ \\
    \midrule
    \multicolumn{10}{@{}l}{\emph{Correlated Gaussian}} \\
    FM & Standard & 4.1$\pm$0.0 & 23.58$\pm$0.01 & 39.6$\pm$0.0 & 0.99$\pm$0.00 & 1.00$\pm$0.00 & 0.58$\pm$0.00 & -- & -- \\
        & Latent   & \textbf{2.9$\pm$0.0} & \textbf{2.27$\pm$0.01} & \textbf{6.8$\pm$3.6} & \textbf{0.45$\pm$0.05} & \textbf{0.83$\pm$0.05} & \textbf{0.51$\pm$0.00} & -- & $1.8\times$ \\
    DM & Standard & 3.6$\pm$0.0 & 16.31$\pm$0.05 & 39.6$\pm$0.0 & 0.97$\pm$0.00 & 1.00$\pm$0.00 & 0.61$\pm$0.01 & -- & -- \\
        & Latent   & \textbf{3.0$\pm$0.0} & \textbf{2.26$\pm$0.00} & \textbf{7.5$\pm$3.5} & \textbf{0.44$\pm$0.05} & \textbf{0.82$\pm$0.05} & \textbf{0.51$\pm$0.00} & -- & $3.6\times$ \\
    NF & Standard & 7.6$\pm$0.1 & 7.00$\pm$0.13 & 50.0$\pm$0.0 & 0.04$\pm$0.00 & 0.09$\pm$0.01 & 0.97$\pm$0.00 & -- & -- \\
        & Latent   & \textbf{2.9$\pm$0.0} & \textbf{2.25$\pm$0.00} & \textbf{6.5$\pm$3.1} & \textbf{0.46$\pm$0.05} & \textbf{0.84$\pm$0.05} & \textbf{0.51$\pm$0.00} & -- & $2.3\times$ \\
    \addlinespace
    \multicolumn{10}{@{}l}{\emph{Fashion-MNIST deblurring}} \\
    FM & Standard & 3.6$\pm$0.0 & \textbf{1.24$\pm$0.01} & 4.6$\pm$0.7 & 0.46$\pm$0.02 & 0.83$\pm$0.02 & \textbf{0.52$\pm$0.00} & \textbf{87.9$\pm$0.2} & -- \\
        & Latent   & \textbf{3.4$\pm$0.0} & 1.27$\pm$0.01 & \textbf{2.3$\pm$0.1} & \textbf{0.51$\pm$0.01} & \textbf{0.88$\pm$0.00} & 0.76$\pm$0.00 & \textbf{87.9$\pm$0.1} & $62\times$ \\
    DM & Standard & 3.5$\pm$0.0 & \textbf{1.19$\pm$0.01} & 5.9$\pm$0.2 & 0.59$\pm$0.00 & \textbf{0.89$\pm$0.00} & \textbf{0.52$\pm$0.00} & \textbf{88.3$\pm$0.2} & -- \\
        & Latent   & \textbf{3.4$\pm$0.0} & 1.25$\pm$0.00 & \textbf{2.2$\pm$0.1} & \textbf{0.50$\pm$0.00} & 0.87$\pm$0.00 & 0.80$\pm$0.00 & \textbf{88.3$\pm$0.1} & $50\times$ \\
    \addlinespace
    \multicolumn{10}{@{}l}{\emph{Gaussian random fields}} \\
    FM & Standard & 11.4$\pm$0.0 & \textbf{6.23$\pm$0.00} & \textbf{1.0$\pm$0.1} & \textbf{0.50$\pm$0.00} & \textbf{0.88$\pm$0.00} & \textbf{0.50$\pm$0.00} & 1.9$\pm$0.0 & -- \\
        & Latent   & 11.5$\pm$0.0 & 6.24$\pm$0.01 & 2.3$\pm$0.2 & 0.48$\pm$0.00 & 0.87$\pm$0.00 & 1.00$\pm$0.00 & 2.0$\pm$0.0 & $42\times$ \\
    DM & Standard & 11.4$\pm$0.0 & \textbf{6.24$\pm$0.01} & \textbf{2.0$\pm$0.6} & \textbf{0.49$\pm$0.01} & \textbf{0.87$\pm$0.01} & \textbf{0.51$\pm$0.00} & 1.9$\pm$0.0 & -- \\
        & Latent   & 11.4$\pm$0.0 & \textbf{6.24$\pm$0.00} & 3.1$\pm$0.2 & 0.47$\pm$0.00 & 0.86$\pm$0.00 & 1.00$\pm$0.00 & 2.0$\pm$0.0 & $19\times$ \\
    \addlinespace
    \multicolumn{10}{@{}l}{\emph{Map to satellite inference}} \\
    DM & Standard & 18.6 & 11.36 & 30.9 & 0.19 & 0.45 & 0.93 & 18.7 & -- \\
        & Latent   & 14.9 & 8.24 & 5.7 & 0.44 & 0.82 & 0.98 & 27.3 & $19\times$ \\
    \bottomrule
  \end{tabular}

  \smallskip
  \raggedright
  \textsuperscript{$\ast$}Coverage is optimal at $0.5$ for Cov$_{50}$ and $0.9$ for Cov$_{90}$. Lower means the posterior is too narrow, higher too wide.

  \textsuperscript{$\dagger$}C2ST accuracy is optimal at $0.5$, where draws and true targets are indistinguishable. Higher means easier to tell apart.

  \textsuperscript{$\ddagger$}NRMSE, CRPS, calibration error, and SSIM are reported in units of $10^{-2}$.
\end{table}

Within each generative family, we compare latent and standard (target-space) inference on posterior draws and on the metrics of \autoref{sec:metrics}, summarized in \autoref{tab:results}. Latent draws inherit the limitations of the autoencoder, which we summarize in Appendix~\ref{app:ae-caveats}.

\paragraph{Correlated Gaussian} Both variants recover the block variances (\autoref{fig:results-toy}), but the latent variant is better calibrated and closer to the exact posterior (\autoref{fig:results-toy-exact}). The quantitative metrics confirm this. The latent variant achieves lower NRMSE and CRPS, a lower calibration error, coverage closer to the nominal levels, and a C2ST accuracy closer to chance.

\begin{figure}[t]
    \centering
    \includegraphics[width=\linewidth]{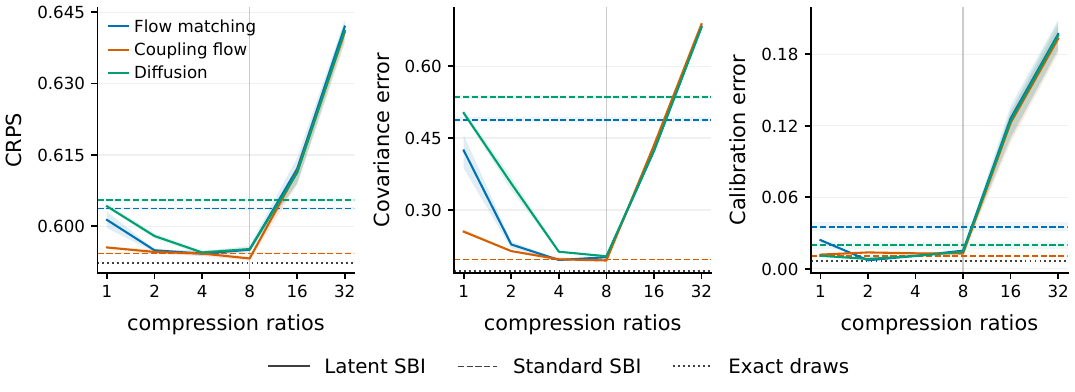}
    \caption{\textbf{Compression ratio study for the Correlated Gaussian example.} Given latent dimension $d$, the latent compression ratios are $64/d$. For each ratio, we evaluate the continuous ranked probability score (CRPS; \textit{left}), covariance error (\textit{middle}), and calibration error (\textit{right}) against their standard SBI baselines for three inference networks: flow matching, normalizing flow, and diffusion model.}
    \label{fig:toy-example-compressions}
\end{figure}

The normalizing flow benefits most from latent inference. In target space it performs poorly, with far too narrow posteriors and a C2ST accuracy close to 1, while in latent space it performs on par with flow matching and diffusion. On this problem, latent inference thus makes the normalizing flow a competitive fast inference network. Latent sampling is \speedupToyFM$\times$ to \speedupToyDM$\times$ faster, including \speedupToyCF$\times$ for the normalizing flow, which samples in a single network pass.

We also studied the effect of varying compression ratio on inferential quality. As shown in \autoref{fig:toy-example-compressions}, at low compression ratios, accuracy and error metrics are at or below the thresholds established by the standard SBI baseline across all three inference networks we studied. However, beyond an optimal compression ratio ($d=8$), information loss becomes dominant, and the inferential quality quickly degrades.

\paragraph{Fashion-MNIST deblurring} Draws from both variants look convincing (\autoref{fig:results-fashion}). Both variants score similarly in NRMSE and CRPS, while the latent variant has lower calibration error and similar coverage. The C2ST favors standard inference.

\paragraph{Gaussian random fields} The posterior is not conditioned on the noise the simulator draws. We therefore expect draws to match only the amplitude and smoothness of the field, not its pixel values. Draws from both variants visually meet this expectation (\autoref{fig:results-fields}). The C2ST nevertheless separates latent draws from true fields, since they lack the highest frequencies (Appendix~\ref{app:ae-caveats}). NRMSE and SSIM are pixel-based, so even exact posterior draws would score far from their optimal values here. We report them for completeness.

\paragraph{Map to satellite} Latent draws follow the street layout of the map, while standard draws mostly show generic vegetation (\autoref{fig:results-map2sat}). The latent variant scores better on every accuracy and marginal calibration metric and attains a higher SSIM. The C2ST separates the draws of both variants from true images.

\paragraph{Joint calibration} Where the C2ST separates latent draws from true targets, the draws largely lack the high frequencies that the autoencoder does not reconstruct, so the loss of joint calibration stems mostly from the manifold of the autoencoder rather than from the posterior (Appendix~\ref{app:ae-caveats}). We present our systematic ablation results in \autoref{app:ablations}.

\section{Discussion and Conclusion}

In this work, we characterized the conditions under which latent-space simulation-based inference recovers the desired target posterior, bounded its error by a posterior, a code, and a decoder term, and systematically analyzed its empirical trade-offs across four case studies and three generative families. Qualitatively and quantitatively, latent SBI achieved similar or better accuracy and marginal calibration than standard inference with considerably faster sampling. As amortized SBI increasingly turns toward inference over very high-dimensional parameter spaces, these results suggest that our two-stage framework is a competitive choice that improves sampling speed while retaining or increasing the quality of the posterior.

\paragraph{Limitations and outlook} Our autoencoder is trained on the parameters alone, so, in practice, the code can discard directions of $\btheta$ that the observation constrains. Therefore, codes conditioned on the observation could be advantageous. Latent draws are also confined to the manifold of the autoencoder. As a result, our pixel-wise decoder loses high-frequency detail that the C2ST metric reliably detects (Appendix~\ref{app:ae-caveats}). Perceptual or adversarial losses could reduce this gap. 

All comparisons are made at a single training budget, and changes to the relative performance of latent and standard inference with the budget remain to be studied. We consider compression only of dense vectors and images, but high-dimensional targets in SBI also arise as sequences, such as time-varying parameters or latent state trajectories \citep{schumacher2023superstatistics, khabibullin2022state}, or as graphs, such as causal structures, hierarchical models, or graph data \citep{lorch2022avici, habermann2025multilevel, jedhoff2026graphs}. These structures call for different inductive biases in the autoencoder and remain unexplored here. Finally, since the autoencoder depends only on the prior over $\btheta$, it could be shared across simulators and tasks with the same prior.

\newpage
\subsection*{AI use statement}
In this work, we used generative AI tools to help develop the theoretical framework, formulate mathematical claims, provide ingredients for and assist in writing their proofs, propose and refine hypotheses, design and provide feedback on the experimental methodology, implement methods, and interpret results. We have \textbf{not} used generative AI tools to generate synthetic data sets or to create any conceptual visualization (especially \autoref{fig:architecture}). Dataset processing and qualitative analysis are not applicable to this work. Additionally, we used generative AI tools to draft and edit parts of the paper, create and edit code, summarize and identify literature, and format references. We have reviewed all AI-assisted work: every mathematical claim and proof was re-derived by an author, AI-written code was reviewed and tested by an author, every citation was verified against a bibliographic database, and all AI-drafted text was revised by the authors. We take responsibility for the final content of this work, including text, claims or artifacts produced with the aid of generative AI.

\subsection*{Reproducibility statement}
 Additional details about the experimental setup and reproducibility requirements are in Appendices~\ref{app:simulators}--\ref{app:compute}.

\ificlrfinal
\subsubsection*{Acknowledgments}
LK was funded by the Deutsche Forschungsgemeinschaft (DFG, German Research Foundation) under Project 528702768.
STR and JMH were funded by the National Science Foundation under Grant No.~2448380.
PCB was funded by the DFG Collaborative Research Center 391 (Spatio-Temporal Statistics for the Transition of Energy and Transport) -- 520388526. This work has been partly supported by the Research Center Trustworthy Data Science and Security (\url{https://rc-trust.ai}), one of the Research Alliance centers within the UA Ruhr (\url{https://uaruhr.de}).
\fi
\newpage
\bibliography{refs} \bibliographystyle{iclr2027_conference}

\clearpage \appendix

\section{Case Study Details}
\label{app:simulators}

\begin{table}[h]
    \centering
    \caption{\textbf{Overview of the case studies.} Parameter dimension denotes the
    dimensionality of the inference target $\btheta$, condition dimension that
    of the conditioning variable, and simulation budget the number of training examples.}
    \label{tab:case-study-overview}

    \resizebox{\textwidth}{!}{%
    \begin{tabular}{lrrr}
        \toprule
        \textbf{Case study}
        & \textbf{Parameter dimension}
        & \textbf{Condition dimension}
        & \textbf{Simulation budget} \\
        \midrule
        Correlated Gaussian
        & $2{,}080$
        & $64$
        & $10{,}000$ \\

        Fashion-MNIST deblurring
        & $32 \times 32 = 1{,}024$
        & $32 \times 32 = 1{,}024$
        & $60{,}000$ \\

        Gaussian random fields
        & $32 \times 32 = 1{,}024$
        & $2$
        & $60{,}000$ \\

        Map to satellite
        & $256 \times 256 \times 3 = 196{,}608$
        & $256 \times 256 \times 3 = 196{,}608$
        & $40{,}555$ \\
        \bottomrule
    \end{tabular}%
    }
\end{table}

\paragraph{Correlated Gaussian.} The $D = 64$ coordinates are split into $k = 8$ contiguous
blocks of equal size. Block $j$ has scale $\sigma_j$ with $\log\sigma_j \sim \mathcal{N}(0, 0.3^2)$,
and the covariance is $\Sigma = \operatorname{blockdiag}_j\, \sigma_j^2 \mathbf{1}\mathbf{1}^\top$, so the
coordinates within a block are perfectly correlated. The target is the upper triangle of
$\Sigma$, $D(D+1)/2 = 2{,}080$ entries, of which $288$ are nonzero. An observation consists of
$n = 32$ draws $\mathbf{y} \sim \mathcal{N}(\mathbf{0}, \Sigma)$, summarized by the
per-coordinate mean of $\mathbf{y}^2$, which is sufficient for the scales and enters both
variants through an identity summary network. Given $\bx$, the block second moment $x_j$
satisfies $n x_j / \sigma_j^2 \sim \chi^2_n$, so the posterior factorizes into $k$
one-dimensional densities in $\log\sigma_j$, which we sample exactly by inverse CDF on a grid.
The target depends linearly on $(\sigma_1^2, \dots, \sigma_k^2)$ and therefore lies in a
$k$-dimensional linear subspace. The optimal linear code of any dimension $m \geq k$
reconstructs it exactly, so the analytic reconstruction floor of the $16$-dimensional code is
zero, and the trained autoencoder reaches a held-out reconstruction RMSE of \aeReconToy. For
the recovery and calibration plots and the two-sample test, each draw is reduced to its block
variances, the mean of each block's entries, which is its least-squares projection onto that
subspace. The study trains on $10{,}000$ targets and evaluates on $1{,}000$ ($10{,}000$ for the two-sample test). The dataset is
fixed across seeds, and each seed re-draws the initialization. The study trains on $8$ cores
of one CPU machine and, unlike the image studies, is timed with the adaptive solvers rather than
at a fixed step count (\autoref{app:compute}).

\paragraph{Fashion-MNIST deblurring.} Targets are Fashion-MNIST images scaled to $[0, 1]$, with
$60{,}000$ for training and the first $1{,}000$ of the $10{,}000$ test images for evaluation. The observation is generated in three
steps: Poisson shot noise at the native $28\times28$ resolution with the standard
data-dependent photon count and a gain of $0.5$, a bilinear resize to $32\times32$, and a
Gaussian blur with reflect-padded borders. The blur width is $\sigma = 2.5$ in native pixels
and is rescaled with the resolution, so the kernel covers the same fraction of the image at
any size. The gain applies to the observation only, so observations are dimmer than their
targets.

\paragraph{Gaussian random fields.} Conditional on its parameters the field is a linear map of
Gaussian noise, so $p(\btheta \mid \log\sigma, \alpha)$ is a zero-mean Gaussian on the
$N\times N$ grid with
\begin{equation*}
  \operatorname{Cov}(\theta_i, \theta_j)
  = \frac{1}{N^{4}}\sum_{k} P(k)\,\cos\!\left(\frac{2\pi k \cdot (i-j)}{N}\right),
  \quad P(k) = \sigma^{2}\,(\alpha|k|)^{-\alpha},
\end{equation*}
and $P(0)=0$. The covariance depends on $i-j$ alone, so the process is stationary, and on the
norm of the wrapped frequency alone, so it is isotropic up to the symmetries of the square
lattice. FFT synthesis makes the covariance circulant, so the field is periodic. Removing the
zero mode leaves the constants as an exact null direction: every realization lies in
$\{\sum_i \theta_i = 0\}$, the conditional has rank $N^2-1$, and analytic comparisons are made
on that subspace. For $\alpha \geq 2$, about $98\,\%$ of the prior mass, the spectrum is not
integrable at the origin in two dimensions, so the continuum field is defined only up to an
additive constant, with stationary increments of index $H = (\alpha - 2)/2$. The grid supplies
the infrared cutoff; the field scale therefore depends on the resolution, and the NRMSE
normalization is the value range of the first $32$ test fields. At small scales the field has the regularity of
a Mat\'ern field with $\nu = (\alpha - 2)/2$, which is near $1/2$ at the prior center
$\alpha = 3$: continuous but not differentiable. The scale $\sigma$ is not the marginal
standard deviation, since the per-sample frequency rescaling below contributes a factor
$\alpha^{-\alpha}$. Because the autoencoder is nonlinear, the latent-space conditional is not
Gaussian, although the target-space conditional is.

The priors are $\log\sigma \sim \mathcal{N}(0, 0.3^2)$ and $\alpha \sim \mathcal{N}(3.0, 0.5^2)$.
A field is drawn by multiplying complex white noise by $\sqrt{P(k)}$ on the FFT frequency grid
and taking the real part of the inverse transform, with the zero mode removed so that every
field has zero sample mean. The frequency grid is scaled per sample by $|\alpha|$; on a discrete
grid this multiplies $P$ by a constant and leaves the correlation structure unchanged, and it
narrows the spread of field magnitudes across the prior.

\paragraph{Map to satellite.} The corpus \citep{sun2025cscmg} provides pixel-aligned
$256\times256$ map and satellite tiles at five zoom levels, from two cities (Glasgow and London),
sourced from a commercial tile service. We use zoom level 17 with the dataset's own split, $40{,}555$ training
and $300$ test pairs. Residual misalignment between map and image comes from differing
acquisition dates of the imagery and the map's vector data. Tiles are used at their native
resolution, stored as 8-bit, scaled to $[0, 1]$, and not augmented. The autoencoder maps
$256\times256\times3$ to a $16\times16\times8$ code, a $96{:}1$ compression against $2{:}1$ on
the $32\times32$ studies. The map tile enters the latent network through a residual encoder
onto the code grid and the target-space network by channel concatenation. The study uses diffusion only.

\section{Model Configurations and Hyperparameters}
\label{app:config}

\begin{table}[t]
  \centering
  \caption{\textbf{Network architecture per case study and space (\autoref{app:config}).} Subnet
  widths are the stage widths of the U-Net (image studies) or the hidden widths of the MLP or
  normalizing-flow subnet (correlated Gaussian). Dropout $0.1$ and, for the U-Net, group
  normalization with $4$ groups are shared by every variant and every stage. The autoencoder
  column gives the encoder / decoder ResNet filters and the resulting code shape, with the
  compression ratio by value count (and, for the image studies, by spatial grid position). Latent
  generators condition on the summary network's output, concatenated channel-wise on the code
  grid. The correlated Gaussian's conditioning variable is already its sufficient statistic, so both variants use
  the identity summary. Fashion-MNIST deblurring and the random fields share one
  configuration. On the random fields, the two conditioning scalars enter as two constant
  channels, which add a few hundred weights to the first layer (\autoref{tab:training-params}).}
  \label{tab:training-arch}
  \footnotesize
  \setlength{\tabcolsep}{3pt}
  \begin{tabularx}{\textwidth}{@{}l l l X l@{}}
    \toprule
    Study & Space & Subnet widths & Autoencoder (enc.\ / dec.\ filters $\to$ code) & Summary \\
    \midrule
    Fashion-MNIST / & target & UNet [68, 140], 2 stages & --- & Identity \\
    random fields ($32^2{\times}1$) & latent & UNet [64], 1 stage
      & [32,64,128,128] / [128,128,64,32] $\to 4^2{\times}32$ (2:1, 64:1 spatial) & ResNet $\to 4^2{\times}64$ \\
    \addlinespace
    Correlated Gaussian & target & MLP [186, 186] & --- & Identity \\
    ($2{,}080$, $k{=}8$) & latent & MLP [139, 139]\textsuperscript{$\dagger$} & MLP [128,128] / [128,128] $\to 16$ (130:1) & Identity \\
    \addlinespace
    Map $\to$ satellite & target & UNet [32,64,128,256,256], 5 stages & --- & Identity \\
    ($256^2{\times}3$) & latent & UNet [140,280], 2 stages
      & [32,64,128,256] / [256,128,64,32] $\to 16^2{\times}8$ (96:1, 256:1 spatial) & ResNet $\to 16^2{\times}64$ \\
    \bottomrule
  \end{tabularx}
  \\[2pt]
  \footnotesize \textsuperscript{$\dagger$}A 6-layer normalizing flow (correlated
  Gaussian only) uses widths [32,32] in target space and [157,157] on the same code.
\end{table}

\begin{table}[t]
  \centering
  \caption{\textbf{Parameter counts and training budgets.} Trainable and frozen parameter counts, computed from the configured architecture
  (\autoref{tab:training-arch}), and the per-run training budget of
  \autoref{app:protocol}. Latent total $=$ frozen autoencoder $+$ trainable summary
  network $+$ trainable generator, matched to the target-space total within $4\%$ in every
  study. The first stage trains for $1{,}600$ autoencoder epochs on Fashion-MNIST and
  the random fields and $300$ on the correlated Gaussian, a fixed design choice in both cases. Its
  share of the run's budget depends on the measured epoch time (\pretrainShareImage\ on
  Fashion-MNIST and the random fields), except on the map study, where the epoch count is instead
  solved so the first stage takes exactly one third of the budget.}
  \label{tab:training-params}
  \footnotesize
  \setlength{\tabcolsep}{3pt}
  \begin{tabular}{@{}l r rrr r r@{}}
    \toprule
    & Target-space & \multicolumn{3}{c}{Latent} & & Budget \\
    \cmidrule(lr){3-5}
    Study & total & AE (frozen) & Summ. & Gen. & Latent total & (h) \\
    \midrule
    Fashion-MNIST      & 2{,}763{,}426  & 1{,}479{,}215 & 811{,}362 & 566{,}164   & 2{,}856{,}741  & 6 \\
    Random fields      & 2{,}764{,}038  & 1{,}479{,}215 & 811{,}684 & 566{,}164   & 2{,}857{,}063  & 6 \\
    Corr.\ Gaussian    & 944{,}635      & 840{,}458     & 0         & 105{,}566   & 946{,}024      & 0.25\textsuperscript{$\dagger$} \\
    Map $\to$ sat.\     & 15{,}424{,}368 & 3{,}242{,}856 & 959{,}598 & 11{,}157{,}865 & 15{,}360{,}319 & 36 \\
    \bottomrule
  \end{tabular}
  \\[2pt]
  \footnotesize \textsuperscript{$\dagger$}Per run, on $8$ CPU cores rather than the GPU,
  so not comparable with the GPU budgets above it. Normalizing flow (correlated Gaussian only,
  depth $6$): target space $1{,}314{,}803$, latent $840{,}458 + 0 + 470{,}169 = 1{,}310{,}627$.
\end{table}

\textbf{Networks.} All models are implemented in BayesFlow~2 \citep{kuhmichel2026bayesflow} on Keras~3 \citep{chollet2015keras} with the JAX backend \citep{jax2018github}. Both spaces share one subnet family per study: a BayesFlow \texttt{UNet} (dropout $0.1$, $4$ groups, $1$ residual block per stage) for the three image-shaped targets, and either an MLP time-embedded subnet (BayesFlow's default \texttt{FlowMatching} / \texttt{DiffusionModel} subnet) or a $6$-layer normalizing flow built from affine coupling layers for the $2{,}080$-dimensional target of the correlated Gaussian. Widths, code shapes and parameter counts are in \autoref{tab:training-arch} and \autoref{tab:training-params}. In every study the latent variant's \emph{total} count, frozen autoencoder included, is matched to the target-space total within $4\%$. Fashion-MNIST deblurring and the random fields share one model configuration. On the random fields, the two conditioning scalars enter as two constant channels, which add a few hundred weights to the first layer, so their parameter counts differ slightly (\autoref{tab:training-params}).

\textbf{Autoencoders.} Every latent variant is preceded by a variational autoencoder (convolutional on the image studies, MLP on the correlated Gaussian), pretrained and frozen before the amortized inference network trains (\autoref{app:protocol}). The latter models
codes standardized per channel, with a mean and standard deviation estimated once after the first stage
and held fixed, and sampling inverts this standardization before decoding. All four use the same InfoVAE convention \citep{zhao2019infovae}, $\alpha = 1 - 10^{-6}$ and $\lambda = 10^{-6}$, giving a KL weight of $10^{-6}$ and no maximum mean discrepancy (MMD) term (\autoref{app:proofs} relates this to the first-stage objective). The encoder's log-variance is clipped to $[-20, 5]$. Every latent posterior on the three image studies conditions through a learned summary network, a residual encoder producing a grid matching the code; every target-space variant instead uses the identity summary, concatenating a pixel-aligned observation directly where one exists, as does the correlated Gaussian's latent variant, whose vector conditioning already is the sufficient statistic and needs no grid.

\textbf{Optimization.} Shared across every study and both stages: AdamW, weight decay $0.01$, per-tensor gradient-norm clipping at $1.5$, dropout $0.1$ on every subnet, batch size $128$ throughout, including the correlated Gaussian. The learning-rate schedule is one-cycle cosine over all training steps, warming from $1.2\times10^{-5}$ to a peak of $3\times10^{-4}$ over the first $30\%$ of steps, then annealing to about $10^{-7}$. The first stage uses the same shape and peak and anneals to about $3\times10^{-8}$. Seeds $0$--$2$ for every variant except the map study, run at seed $0$ only. Numerical precision and hardware are described in \autoref{app:compute}.

\textbf{Budgets and epochs.} Fashion-MNIST and the random fields train for $6$\,h per run. The training-set size ablation scales this to $6\cdot N/60000$\,h at equal epochs. The first stage trains for $1{,}600$ autoencoder epochs on Fashion-MNIST and the random fields, a fixed design choice, and $300$ on the correlated Gaussian, where a vector autoencoder this size converges within a few hundred epochs. Unlike the epoch counts themselves, the \emph{share} of budget they consume depends on the measured epoch time (\pretrainShareImage\ on Fashion-MNIST and the random fields). The correlated Gaussian trains for $0.25$\,h per run on $8$ CPU cores, and the map study for $36$\,h per variant, with the first stage fixed at exactly one third of the budget at its measured autoencoder step time, so its epoch count follows from that fraction rather than a fixed number. The number of epochs is the remaining budget divided by the measured time per epoch (\autoref{app:compute}), rounded up. For the flow-matching pair of Fashion-MNIST at the $6$\,h budget, \stepMsData\ (\stepMsLatent) ms per step give \epochsData\ (\epochsLatent) epochs for the target-space (latent) variant. During training, each run is monitored on a single validation batch with $16$ posterior draws. The reported accuracy and calibration values instead come from a separate evaluation at the end of training, with $1{,}000\times64$ draws on Fashion-MNIST, the random fields and the correlated Gaussian and $300\times64$ on map to satellite.

\section{The Equal-Compute Protocol in Detail}
\label{app:protocol}

\paragraph{Training budget.} The budget counts training time only, accumulated per epoch with the timer stopped during validation. Data generation and evaluation are excluded. The autoencoder's recorded training time is added as an offset. Each run stops at the epoch count derived from the measured step time (\autoref{app:config}), so the budget is fixed before training and does not depend on the hardware it runs on. The accumulated time serves as a check.

\paragraph{Wall-clock rather than FLOPs.} On the deblurring study one evaluation of the latent network costs more than two orders of magnitude fewer FLOPs than one of the target-space network (\flopEvalLatent\ against \flopEvalData\ GFLOP per draw), but the small latent grid reaches only half of the device throughput of the full grid, so its wall-clock time per evaluation falls by a much smaller factor than its FLOPs. Matching FLOPs would give the latent variant more optimizer steps than the same hardware completes in the same time.

\section{Calibration Diagnostics}
\label{app:calibration}
\paragraph{Marginal coverage.} A latent variant's draws lie in the range of the decoder, a set
of lower dimension than the target space and hence of zero volume under the true posterior.
Every per-element marginal of such draws can still be calibrated, so marginal coverage cannot
detect this.

\paragraph{Classifier two-sample test.} Class A pairs each test observation with its true
target, and class B pairs the same observation with one posterior draw. The two classes share
the marginal over observations, so a classifier can separate them only through the
conditional, and their joint distributions coincide exactly when the posterior is correct.
Folds are split by condition, early stopping uses a fold that is never scored, and we report
the held-out accuracy. On the image studies the classifier is a small convolutional network that
reads the observation and the draw stacked as channels. On the correlated Gaussian, the
structural zeros and repeated entries of the full target let a classifier separate any
continuous draw trivially, so the test runs on the block variances of each draw
(\autoref{app:simulators}) with a multilayer perceptron on $10{,}000$ held-out conditions.
Exact posterior draws score \cstToyExact\ on this test.

\paragraph{Controls.} Replacing the draws with reconstructions $\dec(\enc(\btheta))$ of the true
target measures how detectable the decoder's range is on its own. Running the test in latent
space, on pairs $(\bx, \enc(\btheta))$ against $(\bx, \latent)$ with
$\latent \sim q_\phi(\latent \mid \bx)$, scores the posterior without the decoder.

\paragraph{Quantization.} On the image studies, both classes and the reconstruction control are
clipped to the image range and rounded to 8-bit levels before the test.

\paragraph{Closed-form covariance.} For the random fields the conditional covariance is known
analytically (\autoref{app:simulators}), and we report the relative Frobenius error of the
sampled covariance against it. With flow matching (diffusion), the error is \covErrFieldsFM\
(\covErrFieldsDM) for standard and \covErrFieldsLatFM\ (\covErrFieldsLatDM) for latent inference,
against a floor of \covErrFieldsFloor\ for the same number of exact draws. The error is dominated
by the low frequencies, which carry most of the variance, and hides the high-frequency deficit of
the latent draws (Appendix~\ref{app:ae-caveats}).

\paragraph{Draws per condition.} The two-sample test uses one draw per condition; coverage and
CRPS use $64$, and the exact-posterior comparison of \autoref{fig:results-toy-exact} uses $100$. The covariance check uses $4{,}096$ draws for each of $16$
conditions, and the same statistic on $4{,}096$ exact draws gives its floor.

\paragraph{Information terms.} \autoref{tab:information} reports the terms of \autoref{prop:decomposition} after training, per study and averaged over seeds. $H(\latent \mid \btheta)$ is closed form from the encoder's log-variances. $H(\latent)$ is the held-out negative log-likelihood of an unconditional normalizing flow fitted to one encoded draw per training target, a cross-entropy and hence an upper bound, which $\MI(\btheta; \latent) = H(\latent) - H(\latent \mid \btheta)$ inherits. The decoder term is the reconstruction negative log-likelihood under $\mathcal{N}(\dec(\latent), \sigma^2 I)$ with $\sigma^2$ set to the held-out mean squared error, minus $H(\btheta \mid \latent) = H(\btheta) - \MI(\btheta; \latent)$; $H(\btheta)$ is available by Monte Carlo over the prior mixture on the random fields, so the term is reported there. On the deblurring study the reconstruction likelihood stands alone, and on the correlated Gaussian the target lies on an $8$-dimensional subspace and has no density. The posterior's density on encoded test targets equals the posterior term up to $H(\latent \mid \bx)$ (normalizing flows exact, flow matching by the instantaneous change of variables; diffusion models expose no likelihood in our implementation), and the latent-space two-sample test (\autoref{tab:information}) reads the same term without a density. The code term of a trained encoder, $\MI(\btheta; \bx) - \MI(\latent; \bx)$, would need a lower bound on $H(\latent)$ or a tight bound on $H(\latent \mid \bx)$, and the difference of the two upper bounds available here is neither a lower bound nor a tight estimate (on the random fields it exceeds $\MI(\btheta; \bx)$), so the table carries the closed-form code term of the reconstruction-optimal linear code of the same dimension instead, next to the closed-form $\MI(\btheta; \bx)$. Two readings follow. In \autoref{sec:theory}, $\MI(\btheta; \latent) = H(\latent) - H(\latent \mid \btheta)$ balances the reconstruction error, which drives the encoder's variance down and $H(\latent \mid \btheta)$ with it, against the KL penalty, which holds the variance up. With the KL weight of $10^{-6}$ used here (\autoref{app:config}) the balance lands far on the reconstruction side: on the correlated Gaussian the code's entropy is \hCodeToyKeight\ nats against a conditional entropy of \hCodeGivenThetaToyKeight\ nats, so most of $\MI(\btheta; \latent)$ measures how far the encoder's variance has collapsed rather than what the code retains of $\btheta$; the decoder term, which carries the same bound, is a loose number for the same reason, and the reconstruction error of each autoencoder (\autoref{app:simulators}) is the quantity that shows what the autoencoder costs. The latent-space two-sample test sits at its null on the correlated Gaussian and on the random fields, and just above it on the deblurring study, indicating that the posterior networks reproduce the encoded conditional. \begin{table}[t]
  \centering
  \caption{\textbf{Information terms of the error decomposition.} Information terms of \autoref{prop:decomposition} after training, in nats, mean
  $\pm$ standard deviation over three seeds (omitted where it rounds to zero). Code
  entropies are taken in the standardized code space the generator models;
  $\MI(\btheta; \latent)$ is invariant to that map. $H(\latent)$ is the held-out
  cross-entropy of an unconditional normalizing flow fitted to encoded training codes, an upper
  bound, so $\MI(\btheta; \latent)$ and the decoder term are upper bounds. The decoder
  term is the reconstruction negative log-likelihood under $\mathcal{N}(\dec(\latent),
  \sigma^2 I)$ with $\sigma^2$ the held-out mean squared error, minus $H(\btheta)$, plus
  $\MI(\btheta; \latent)$; on the deblurring study $H(\btheta)$ is unknown and the
  reconstruction likelihood stands alone. The linear reference is the reconstruction-optimal
  linear code of the same dimension, whose code term is closed form. Generator likelihoods
  are on encoded test targets (normalizing flows exact, flow matching by the instantaneous
  change of variables); diffusion models expose none. The latent-space two-sample test
  separates $(\bx, \enc(\btheta))$ from $(\bx, \latent \sim q_\phi(\latent \mid \bx))$
  pairs, one draw per condition, against a label-permuted null.}
  \label{tab:information}
  \scriptsize
  \setlength{\tabcolsep}{4pt}
  \begin{tabular}{@{}l rrr@{}}
    \toprule
    & correlated Gaussian & deblurring & random fields \\
    \midrule
    $d_{\latent}$ & \codeDimToyKeight & \codeDimFashion & \codeDimFields \\
    $H(\latent \mid \btheta)$ & \hCodeGivenThetaToyKeight & \hCodeGivenThetaFashion & \hCodeGivenThetaFields \\
    $H(\latent)$ (bound) & \hCodeToyKeight & \hCodeFashion & \hCodeFields \\
    $\MI(\btheta; \latent)$ (bound) & \infoCodeToyKeight & \infoCodeFashion & \infoCodeFields \\
    \quad per code dimension & \infoCodePerDimToyKeight & \infoCodePerDimFashion & \infoCodePerDimFields \\
    \midrule
    held-out MSE, $\sigma^2$ & \aeMseToyKeight & \aeMseFashion & \aeMseFields \\
    reconstruction NLL & \reconNllToyKeight & \reconNllFashion & \reconNllFields \\
    $H(\btheta)$ & \hThetaToyKeight & \hThetaFashion & \hThetaFields \\
    decoder term (bound) & \decoderTermToyKeight & \decoderTermFashion & \decoderTermFields \\
    \midrule
    $\MI(\btheta; \bx)$ & \infoThetaXToyKeight & \infoThetaXFashion & \infoThetaXFields \\
    code term, linear reference & \codeTermLinearToyKeight & \codeTermLinearFashion & \codeTermLinearFields \\
    \quad kept by the linear code & \linearKeptToyKeight & \linearKeptFashion & \linearKeptFields \\
    \midrule
    posterior NLL, NF & \genNllCFToyKeight & \genNllCFFashion & \genNllCFFields \\
    posterior NLL, FM & \genNllFMToyKeight & \genNllFMFashion & \genNllFMFields \\
    \midrule
    latent-space C2ST, NF & \cstCodeCFToyKeight & \cstCodeCFFashion & \cstCodeCFFields \\
    latent-space C2ST, FM & \cstCodeFMToyKeight & \cstCodeFMFashion & \cstCodeFMFields \\
    latent-space C2ST, DM & \cstCodeDMToyKeight & \cstCodeDMFashion & \cstCodeDMFields \\
    \quad null band, mean & \cstCodeNullToyKeight & \cstCodeNullFashion & \cstCodeNullFields \\
    \bottomrule
  \end{tabular}
\end{table}

\subsection{Caveats of the Autoencoder}
\label{app:ae-caveats}
Our autoencoders are VAEs trained with a pixel-wise reconstruction loss and a small KL penalty.
This objective favors the directions of largest prior variance (\autoref{sec:theory}) and is
known to lose fine detail in images \citep{larsen2016autoencoding}, and the latent draws inherit
whatever the autoencoder does not reconstruct. On the random fields, the latent draws keep only
\specKeepHiFieldsLatFM\,\% of the analytic power above wavenumber \specCutFields, a band that holds
\specShareHiFields\,\% of the variance. The covariance error barely registers this, but the relative
error of the radially averaged spectrum is \specErrFieldsLat\ for latent against
\specErrFieldsData\ for standard inference (floor \specErrFieldsFloor), and the C2ST separates
latent draws from true fields almost perfectly (\cnnFieldsLat), as it does the autoencoder's own
reconstructions (\cnnFieldsRecon). On map to satellite the C2ST likewise separates latent draws
(\cnnSatLat) about as well as reconstructions (\cnnSatRecon).

\section{Computational Complexity and Hardware}
\label{app:compute}

\paragraph{Where the speedup comes from.} The cost of a posterior draw is the number of network
evaluations times the cost of one evaluation. Compressing the target changes only the second
factor, so the speedup is bounded by the ratio of per-evaluation costs, not by the compression
ratio. For a convolutional network that ratio is set by the number of grid positions; the
channel count of the code enters only the first layer. The deblurring code,
$4\times4\times32$, has half as many entries as the $32\times32$ target but $64$ times fewer
positions, and with the narrower latent network one evaluation costs more than two orders of magnitude
fewer FLOPs. The decoder adds \decoderFlopShare\,\% to the latent variant's FLOPs per draw.
An $8\times8\times8$ code with the same number of entries would save more than an order of magnitude in FLOPs
at the same widths. The measured speedup is smaller than the FLOP ratio
(\autoref{app:protocol}), and it is the measured one we report.

\paragraph{Amortizing the first stage.} Under our accounting the autoencoder's training
is part of the equal budget, so the latent variant costs no additional training. Counted as
additional, as when an autoencoder is added to an existing target-space pipeline, it is repaid
after about \breakEvenFashionFM\ draws on the deblurring study with flow matching: the
\pretrainShareImage\ of the $6$\,h budget spent on the first stage, divided by the difference between
\msPerKdataFM\ and \msPerKlatFM\,ms per $1000$ draws.

\paragraph{Solver steps.} The cost timing fixes the solver and its step count in both spaces at
$30$ steps, which is $180$ network evaluations per draw for flow matching with a fixed-step
Tsitouras integrator and $60$ for the two-step diffusion sampler. It runs on untrained networks
of the configured shapes, since the cost of a fixed workload depends on shapes only. Speedups
are therefore comparable within a family, and absolute costs are not comparable across
families. The correlated Gaussian is the exception: it is timed on CPU with the adaptive
solvers (\autoref{app:simulators}). The adaptive solvers (a Tsitouras 5(4) solver for flow matching, a two-step adaptive SDE
solver for diffusion) produce all reported posterior draws. They take a data-dependent number of
steps and are not part of the reported speedup.

\paragraph{Sampling batch size.} Both variants of a family are timed at the same sampling batch
size, the largest batch that both sample in one pass: $256$ on the $32\times32$ studies ($128$ for diffusion on
the random fields) and $8$ on the
map study. At this size the target-space network is
already compute-bound and the latent network is not, so the reported ratio is a lower bound on
the ratio of attainable throughputs. The latent network's smaller memory footprint would allow
larger batches, which we do not count.

\paragraph{Precision.} The $32\times32$ studies train and sample in float32. The map study
trains both variants in bfloat16 mixed precision and samples every reported draw in float32.

\paragraph{Hardware.} Every image-study timing uses one NVIDIA L40 (46\,GB) on a compute cluster,
as do all flow-matching and map runs. Some diffusion models trained on other Ada-generation
GPUs at the same epoch counts (\autoref{app:config}). The correlated Gaussian trains and is timed on $8$ cores of
a CPU machine of the same cluster, so its costs are not comparable with the GPU ones. GPU runs did not enforce deterministic operations, so reruns with the same seed
can differ slightly. The software is pinned to Python~3.12, JAX~0.11.0 with its CUDA~12 plugin,
Keras~3.15.1 and BayesFlow~2.0.13, and the L40 machine runs NVIDIA driver 595.71.05.

\paragraph{Timing.} Timings run in a fresh process before any sampling, since timings taken
after sampling include recompilation. Step times are measured in short chunks, and any chunk
during which the device clock throttles is discarded.

\section{Proofs}
\label{app:proofs}

\paragraph{Proof of \autoref{prop:decomposition}.}
Write $q_\xi(\btheta, \latent \mid \bx) = q_\xi(\latent \mid \bx)\,
q_\xi(\btheta \mid \latent, \bx)$ and apply the chain rule of the Kullback--Leibler divergence
\citep[Ch.~2]{cover2006elements} to the left-hand side of \eqref{eq:decomposition}:
\begin{equation*}
\begin{split}
\E_{\bx} \KL\big(q_\xi(\btheta, \latent \mid \bx) \,\|\, q_\phi(\latent \mid \bx)\, q_\psi(\btheta \mid \latent)\big)
&= \E_{\bx} \KL\big(q_\xi(\latent \mid \bx) \,\|\, q_\phi(\latent \mid \bx)\big) \\
&\quad + \E_{\bx, \latent} \KL\big(q_\xi(\btheta \mid \latent, \bx) \,\|\, q_\psi(\btheta \mid \latent)\big).
\end{split}
\end{equation*}
Adding and subtracting $\log q_\xi(\btheta \mid \latent)$ inside the second expectation splits it
into $\E_{\bx, \latent} \KL(q_\xi(\btheta \mid \latent, \bx) \,\|\, q_\xi(\btheta \mid \latent)) = \MI(\btheta;
\bx \mid \latent)$ and $\E_{\latent} \KL(q_\xi(\btheta \mid \latent) \,\|\, q_\psi(\btheta \mid \latent))$; the
second expectation is under $q_\xi(\latent)$ because averaging $q_\xi(\btheta \mid \latent, \bx)$ over
$q_\xi(\bx \mid \latent)$ gives $q_\xi(\btheta \mid \latent)$. Since the encoder depends on $\btheta$ alone,
$\latent$ and $\bx$ are conditionally independent given $\btheta$, so the chain rule of mutual
information gives $\MI(\btheta, \latent; \bx) = \MI(\btheta; \bx) = \MI(\latent; \bx) +
\MI(\btheta; \bx \mid \latent)$, hence $\MI(\btheta; \bx \mid \latent) = \MI(\btheta; \bx) -
\MI(\latent; \bx)$. Finally, $p(\btheta \mid \bx)$ and $q_{\phi, \psi}(\btheta \mid \bx)$ are the
marginals over $\latent$ of the two distributions on the left-hand side, and marginalization
does not increase the divergence.
\qed

\paragraph{Proof of \autoref{prop:objective}.} The InfoVAE objective of \citet{zhao2019infovae}, in their notation, is $\E_{p(\btheta)} \E_{q_\xi(\latent \mid \btheta)} \log q_\psi(\btheta \mid \latent) - (1 - \alpha)\, \E_{p(\btheta)} \KL(q_\xi(\latent \mid \btheta) \,\|\, p(\latent)) - (\alpha + \lambda - 1)\, D(q_\xi(\latent) \,\|\, p(\latent))$. The first-stage objective of \autoref{sec:two-stage} is this objective with the Gaussian decoder of \autoref{sec:theory}, with $D$ the maximum mean discrepancy, $\alpha < 1$ and $\lambda > 0$. Our experiments use $\alpha + \lambda - 1 = 0$ (\autoref{app:config}), the case \citet{zhao2019infovae} identify as the $\beta$-VAE family. Their Proposition~2 assumes continuous spaces and sufficiently flexible families for encoder and decoder, and its proof shows that for any fixed value of $\MI(\btheta; \latent)$ the objective is maximized when $q_\psi(\btheta \mid \latent) = q_\xi(\btheta \mid \latent)$ for every $\latent$ and $q_\xi(\latent) = p(\latent)$; its proof gives the maximal value as $(1 - \beta)\, \MI(\btheta; \latent) - H(p(\btheta))$ with $\beta = 1 - \alpha$. The proof uses $\alpha + \lambda - 1 > 0$; at $\alpha + \lambda - 1 = 0$ the marginal term reduces to $-\beta\, \KL(q_\xi(\latent) \,\|\, p(\latent))$, and $\beta > 0$ alone forces $q_\xi(\latent) = p(\latent)$, so the conclusion is unchanged. The first condition is the vanishing of the decoder term of \eqref{eq:decomposition}, which a Gaussian decoder with fixed variance can attain only if $q_\xi(\btheta \mid \latent)$ is Gaussian with that variance; in practice, it is approached, not attained. The value depends on the encoder only through $\MI(\btheta; \latent)$. \qed

\section{Additional Results}
\label{app:additional}

\begin{figure}[!t]
  \centering
  \includegraphics[width=\linewidth]{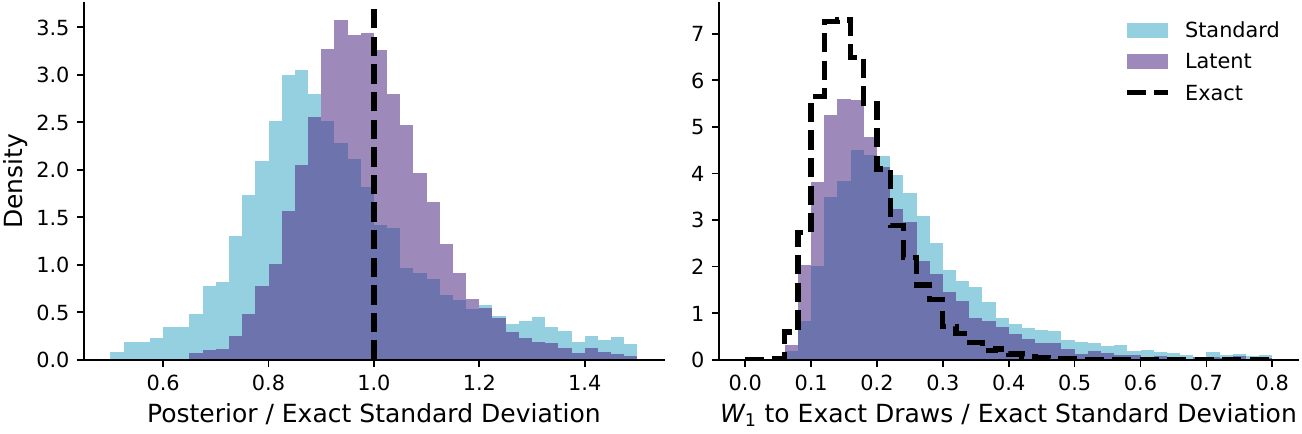}
  \caption{\textbf{Comparison against the exact posterior on the correlated Gaussian.} With flow matching, on the $k = 8$ block variances ($1{,}000$ test conditions, $100$ draws each). \textit{Left:} posterior standard deviation relative to the exact one. Latent widths scatter evenly around the exact width, while standard widths are too narrow for most conditions and far too wide for some. \textit{Right:} Wasserstein-1 distance ($W_1$) to exact draws, relative to the exact standard deviation; the dashed outline shows the sampling noise of the exact posterior itself.}
  \label{fig:results-toy-exact}
\end{figure}

Along the block variances of the correlated Gaussian, standard flow matching is overconfident for most conditions, while latent flow matching balances over- and underconfidence (\autoref{fig:results-toy-exact}).

\section{Ablations}
\label{app:ablations}

\paragraph{Joint fine-tuning of the autoencoder.} Starting from a trained latent flow-matching
model on Fashion-MNIST (seed $0$), we unfroze the encoder and trained encoder and posterior network jointly
for $300$ further epochs, against a control that continued for the same number of epochs with the
encoder frozen. Neither continuation is budget-matched to the main comparison. Joint training was
less accurate than the frozen control, with an NRMSE of \ftNrmseJoint\ against \ftNrmseFrozen\ on
the $1{,}000$ test conditions with $64$ posterior draws each. Because the code standardization is fixed
after the first stage (\autoref{app:config}), the posterior's objective can be lowered by shrinking the
encoder's output. The standard deviation of the standardized code fell to \ftStdJoint\ times its
initial value, while the reconstruction error grew by a factor of only \ftReconJoint. KL weights of
$10^{-3}$ and $10^{-2}$ hold the scale (\ftStdKLmid\ and \ftStdKLhigh\ times its initial value)
but raise the reconstruction error by factors of \ftReconKLmid\ and \ftReconKLhigh, with an NRMSE
of \ftNrmseKLmid\ and \ftNrmseKLhigh.

\paragraph{Training-set size.} We repeat the Fashion-MNIST comparison with $1{,}000$ and
$10{,}000$ training targets instead of $60{,}000$. At each size both variants train for the same
number of epochs as in the main comparison, so the budget scales with the data. Each size trains
its own autoencoder, with a held-out reconstruction RMSE of \aeReconFashionOneK\ at $1{,}000$,
\aeReconFashionTenK\ at $10{,}000$ and \aeReconFashion\ at $60{,}000$ targets.
\autoref{tab:size} reports the metrics at each size.
\begin{table}[t]
  \centering
  \caption{\textbf{Training-set size ablation.} Fashion-MNIST deblurring at three training-set sizes, at the epoch counts of the
  main comparison (\autoref{app:config}). Mean $\pm$ standard error over three seeds.}
  \label{tab:size}
  \scriptsize
  \setlength{\tabcolsep}{3pt}
  \begin{tabular}{@{}lll rrrrr@{}}
    \toprule
    Targets & & & NRMSE $\downarrow$ & CRPS $\downarrow$ & Cal.\ err.\ $\downarrow$ & Cov$_{50}$ & Cov$_{90}$ \\
    \midrule
    $1{,}000$ & FM & Standard & 0.052$\pm$0.000 & 0.0186$\pm$0.0001 & 0.061$\pm$0.020 & 0.58$\pm$0.05 & 0.92$\pm$0.01 \\
     &  & Latent & 0.061$\pm$0.000 & 0.0243$\pm$0.0002 & 0.062$\pm$0.003 & 0.57$\pm$0.01 & 0.86$\pm$0.01 \\
     & DM & Standard & 0.055$\pm$0.000 & 0.0196$\pm$0.0002 & 0.056$\pm$0.005 & 0.54$\pm$0.04 & 0.89$\pm$0.01 \\
     &  & Latent & 0.060$\pm$0.000 & 0.0240$\pm$0.0001 & 0.063$\pm$0.001 & 0.54$\pm$0.01 & 0.85$\pm$0.01 \\
    \addlinespace
    $10{,}000$ & FM & Standard & 0.044$\pm$0.000 & 0.0151$\pm$0.0001 & 0.090$\pm$0.012 & 0.43$\pm$0.06 & 0.84$\pm$0.02 \\
     &  & Latent & 0.048$\pm$0.000 & 0.0178$\pm$0.0001 & 0.041$\pm$0.002 & 0.50$\pm$0.01 & 0.85$\pm$0.01 \\
     & DM & Standard & 0.042$\pm$0.000 & 0.0144$\pm$0.0001 & 0.072$\pm$0.011 & 0.50$\pm$0.06 & 0.86$\pm$0.00 \\
     &  & Latent & 0.045$\pm$0.000 & 0.0167$\pm$0.0001 & 0.042$\pm$0.001 & 0.50$\pm$0.01 & 0.85$\pm$0.00 \\
    \addlinespace
    $60{,}000$ & FM & Standard & 0.036$\pm$0.000 & 0.0124$\pm$0.0001 & 0.046$\pm$0.007 & 0.46$\pm$0.02 & 0.83$\pm$0.02 \\
     &  & Latent & 0.034$\pm$0.000 & 0.0127$\pm$0.0001 & 0.023$\pm$0.001 & 0.51$\pm$0.01 & 0.88$\pm$0.00 \\
     & DM & Standard & 0.035$\pm$0.000 & 0.0119$\pm$0.0001 & 0.059$\pm$0.002 & 0.59$\pm$0.00 & 0.89$\pm$0.00 \\
     &  & Latent & 0.034$\pm$0.000 & 0.0125$\pm$0.0000 & 0.022$\pm$0.001 & 0.50$\pm$0.00 & 0.87$\pm$0.00 \\
    \bottomrule
  \end{tabular}
\end{table}

\section{Additional Random Samples}
\label{app:per-seed}

\begin{figure}[p]
  \centering
  \includegraphics[width=0.9\linewidth]{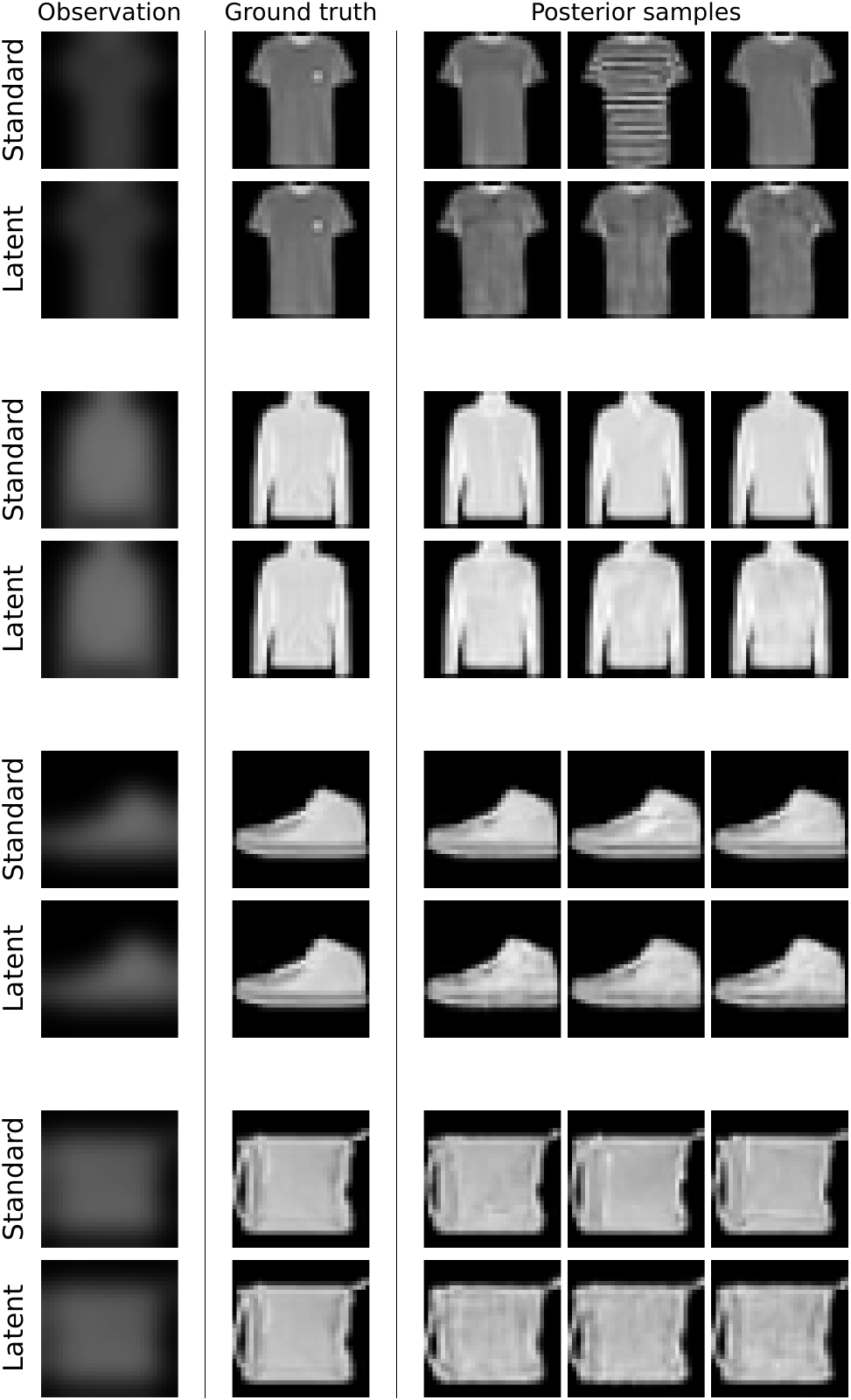}
  \caption{\textbf{Random conditions for Fashion-MNIST deblurring.} Posterior samples for four Fashion-MNIST test conditions drawn at random, not selected,
  with flow matching.}
  \label{fig:random-fashion}
\end{figure}

\begin{figure}[p]
  \centering
  \includegraphics[width=0.9\linewidth]{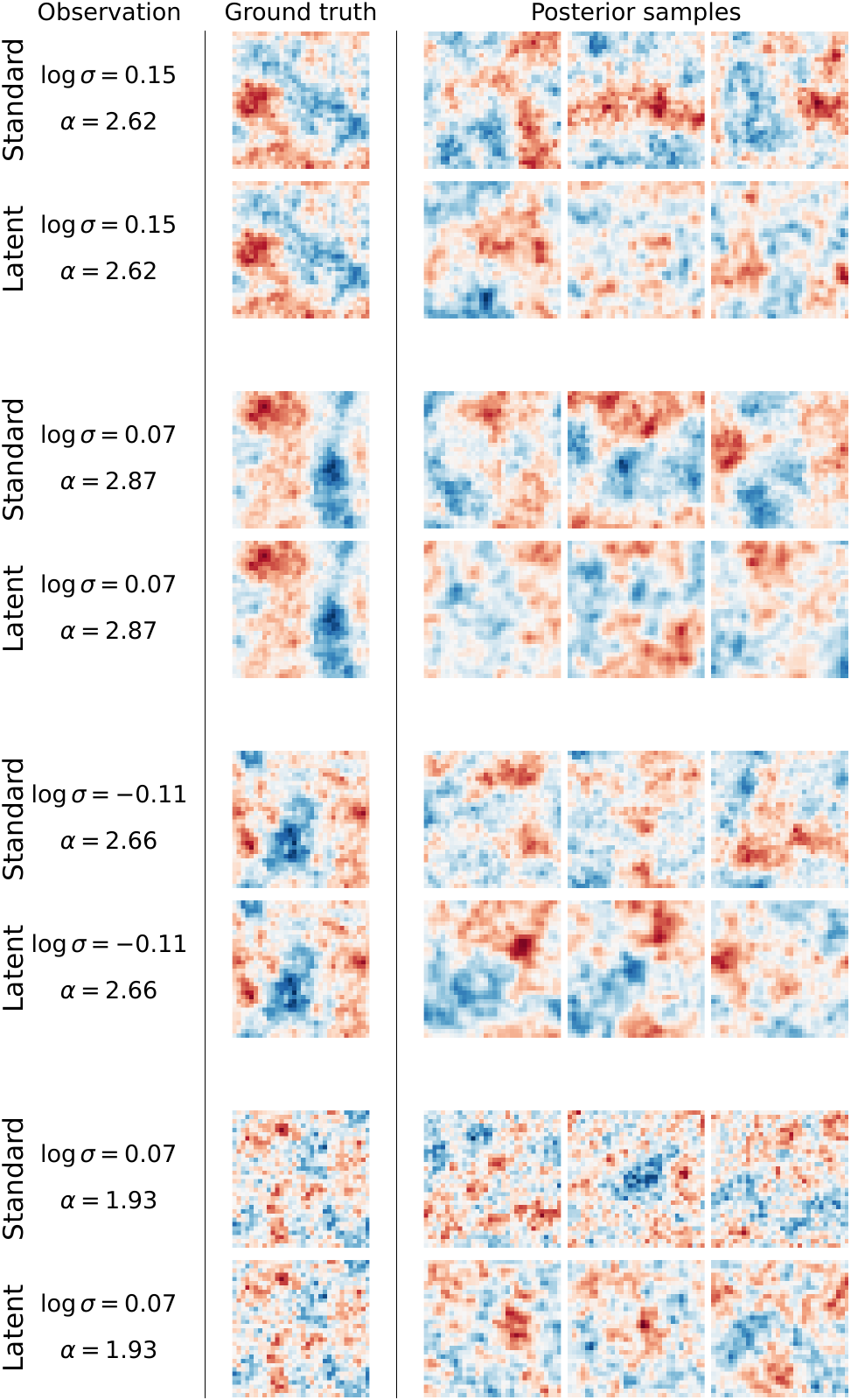}
  \caption{\textbf{Random conditions for the Gaussian random fields.} Posterior samples for four random-field test conditions drawn at random, not selected,
  as in \autoref{fig:results-fields}. Colors share one symmetric scale per condition.}
  \label{fig:random-fields}
\end{figure}

\begin{figure}[p]
  \centering
  \includegraphics[width=0.9\linewidth]{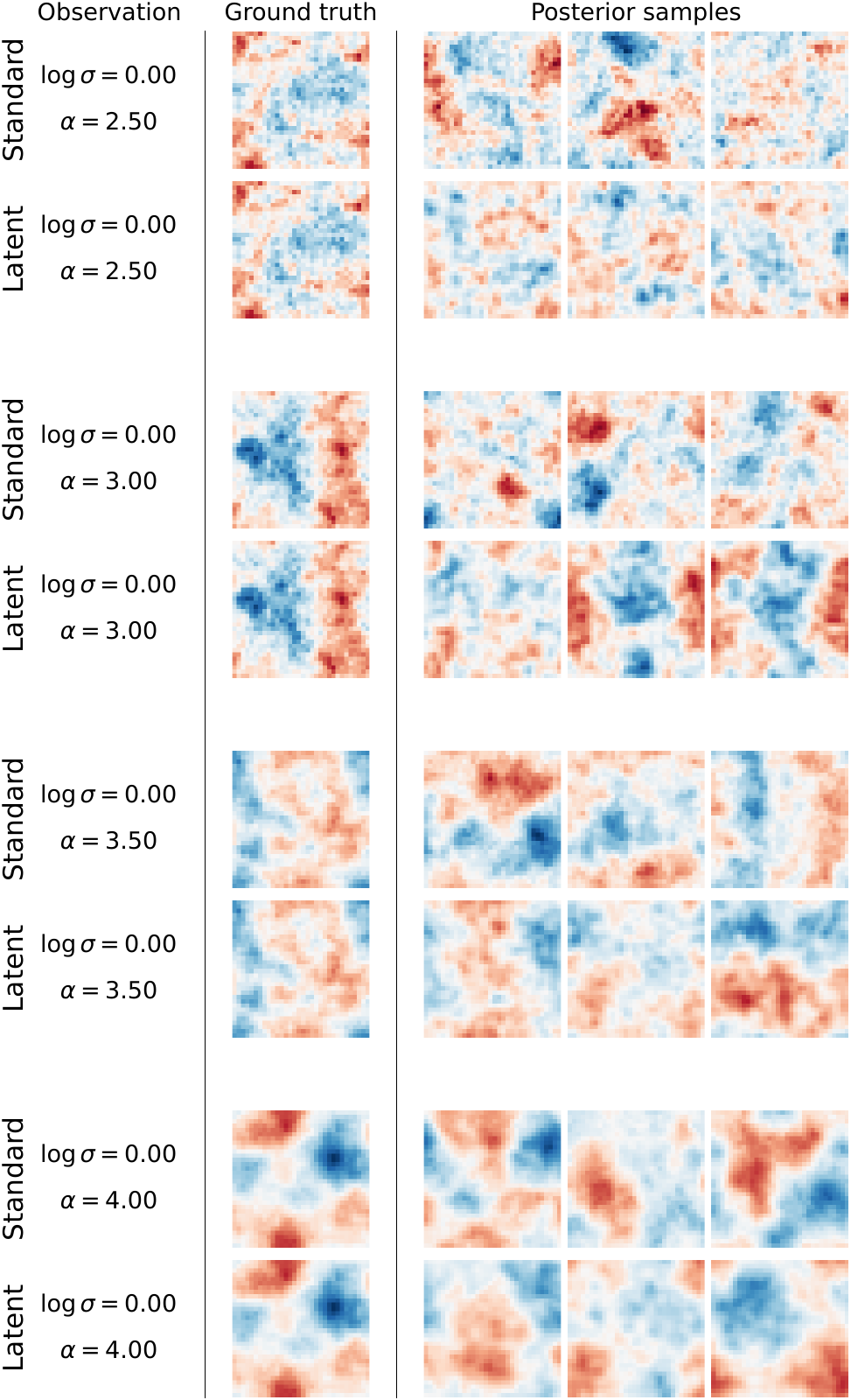}
  \caption{\textbf{Smoothness sweep for the Gaussian random fields.} Posterior samples for the random fields at $\log\sigma = 0$ and four equally spaced values
  of $\alpha$. Smoothness increases with $\alpha$ in both variants, as in the ground truth.}
  \label{fig:fields-alphas}
\end{figure}

\begin{figure}[p]
  \centering
  \includegraphics[width=0.8\linewidth]{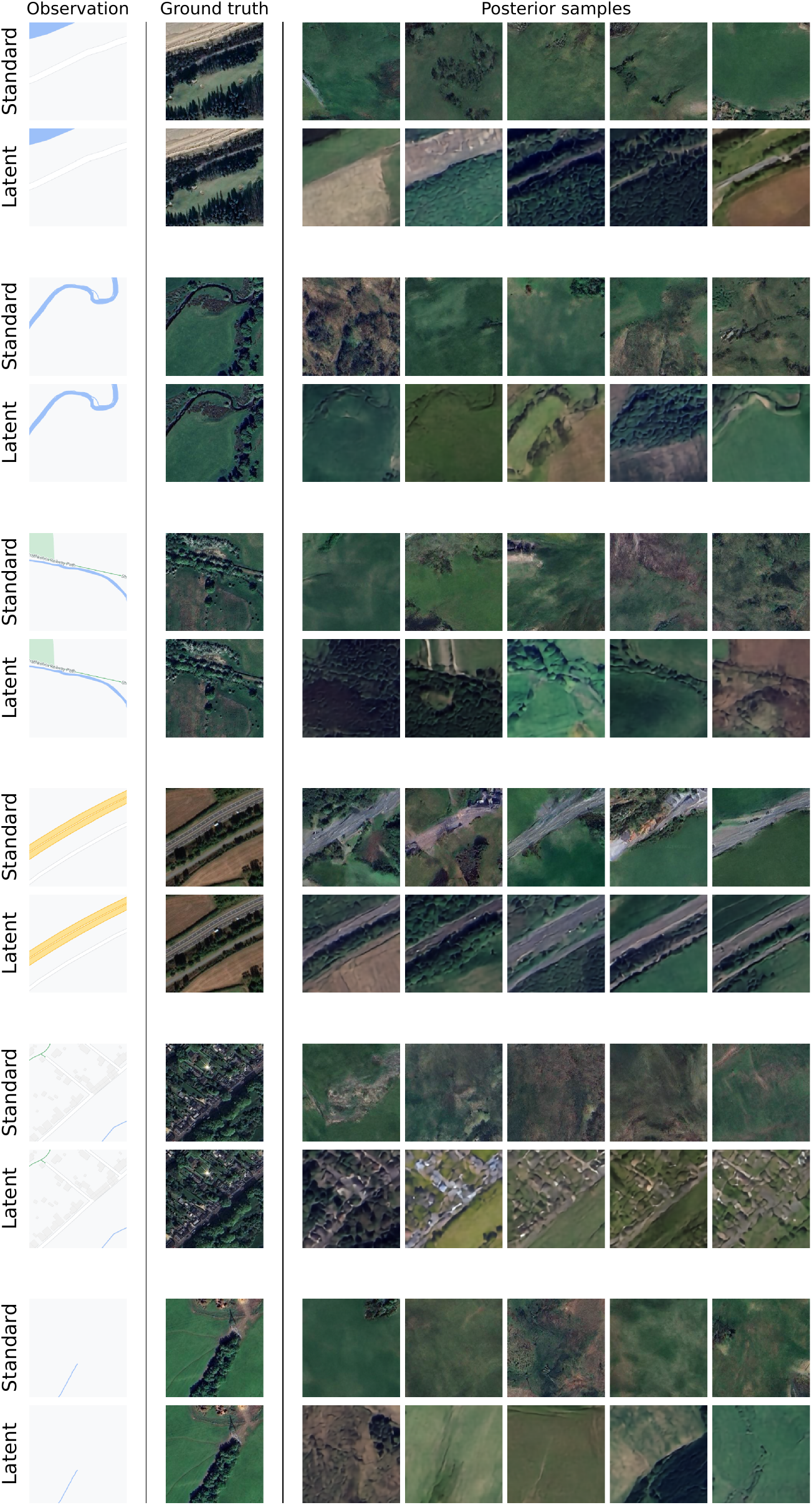}
  \caption{\textbf{Random conditions for map to satellite.} Posterior samples for six map-to-satellite test conditions drawn at random, not
  selected, as in \autoref{fig:results-map2sat}.}
  \label{fig:random-map2sat}
\end{figure}

\end{document}